\documentclass{article}

\usepackage{arxiv}

\usepackage[utf8]{inputenc} % allow utf-8 input
\usepackage[T1]{fontenc}    % use 8-bit T1 fonts
\usepackage{hyperref}       % hyperlinks
\usepackage{url}            % simple URL typesetting
\usepackage{booktabs}       % professional-quality tables
\usepackage{amsfonts}       % blackboard math symbols
\usepackage{nicefrac}       % compact symbols for 1/2, etc.
\usepackage{microtype}      % microtypography
\usepackage{lipsum}
\usepackage{graphicx}
\usepackage{subfigure}
\usepackage{bm}
\usepackage{amssymb}
\usepackage{amsthm}
\usepackage{latexsym}
\usepackage{bbding}
\usepackage{algorithm}
\usepackage{algorithmic}
\usepackage{booktabs}
\usepackage{multirow}
\usepackage{url}
\usepackage{array}
\usepackage{amsmath}
\usepackage{booktabs}
\usepackage{xcolor}
\usepackage{epstopdf}
\usepackage{tabularx}

\title{Preventing Information Leakage with Neural Architecture Search}

\author{
  Shuang Zhang, Liyao Xiang, Congcong Li, Yixuan Wang, Quanshi Zhang\\
  Shanghai Jiao Tong University \\
  \And
  Wei Wang, Bo Li \\
  The Hong Kong University of Science and Technology \\
}

\begin{document}
\maketitle

\begin{abstract}
Powered by machine learning services in the cloud, numerous learning-driven mobile applications are gaining popularity in the market. As deep learning tasks are mostly computation-intensive, it has become a trend to process raw data on devices and send the deep neural network (DNN) features to the cloud, where the features are further processed to return final results. However, there is always unexpected leakage with the release of features, with which an adversary could infer a significant amount of information about the original data. We propose a privacy-preserving reinforcement learning framework on top of the mobile cloud infrastructure from the perspective of DNN structures. The framework aims to learn a policy to modify the base DNNs to prevent information leakage while maintaining high inference accuracy. The policy can also be readily transferred to large-size DNNs to speed up learning. Extensive evaluations on a variety of DNNs have shown that our framework can successfully find privacy-preserving DNN structures to defend different privacy attacks.
\end{abstract}

% keywords can be removed

\section{Introduction}
The landscape of mobile computing has evolved with the recent move of deep learning algorithms from computational backends to the edge, with a number of optimized engines being readily available to application developers. Despite this trend, mobile devices are known for the lack of computational power, and thus not suitable for computation-intensive deep neural network (DNN) operations. 

Instead of storing and processing the entire DNN on the device, a practical approach is to place the computation-intensive parts at the backend cloud.  Such mobile-cloud computational frameworks have been adopted and discussed in \cite{hu2019dynamic, teerapittayanon2017distributed, han2016mcdnn, wang2018not, zhang2016privacy}.  However, it would be a violation of privacy if we place the entire piece of DNN in the cloud and ask the user to upload raw data. Thus it becomes a design choice to select intermediate-layer features instead of the raw data to offload to the cloud. Previous works have demonstrated that unintended input information could be revealed by intermediate-layer representations in \cite{shokri2017membership, ganju2018property,  melis2019exploiting, zhibo2019beyond}, and worse still, attackers can even reconstruct inputs from the accessing features in \cite{dosovitskiy2016inverting, zeiler2014visualizing, mahendran2015understanding}.

As countermeasures, a number of approaches have been proposed, which roughly include two categories: differentially-private perturbation including \cite{wang2018not, shokri2015privacy, abadi2016deep, papernot2016semi} as well as cryptographic methods such as \cite{zhang2016privacy, mohassel2017secureml, mohassel2018aby}. Differentially-private mechanisms have been applied to the model parameters \cite{shokri2015privacy, abadi2016deep}, intermediate features \cite{wang2018not}, model predictions \cite{papernot2016semi}, or objective functions \cite{zhang2012functional}. For example, in \cite{wang2018not}, by nullifying sensitive regions of inputs and injecting random noise to the intermediate features, the device is able to guarantee strong privacy for any perturbed features sent to the cloud. Despite its strong theoretical guarantee, it is unknown how the perturbed features can guard against adversary attacks. The methods rooted in cryptography are either computationally intensive, such as BGV homomorphic encryption schemes \cite{zhang2016privacy}, or proposed under ideal two-party/three-party server assumptions \cite{mohassel2018aby}. Since the mobile device is mostly a thin piece with limited computation power, it is hard to fit the system requirement on storage, latency, power consumption, etc. Other works like \cite{mo2020darknetz} rely on the edge device’s Trusted Execution Environment (TEE) to preserve privacy.

Different from conventional approaches, we propose a mobile-cloud DNN structure tuning framework for seeking a balance among three objectives: accuracy, privacy, and on-device resources costs. Note that privacy in this paper refers to input indistinguishability, and the privacy metric is evaluated as the capability to guard against a generic adversary who tries to infer private input from the released features. We are motivated by the observation that higher-level features are less revealing on the input but incur higher on-device costs. Moreover, we found that by carefully choosing compression techniques, a better tradeoff between privacy and computational complexity can be achieved. Hence, our design space includes the selection of the partition layer between the mobile and the cloud, as well as the combination of different compression techniques. The problem is formulated as an optimization one that searches a distributed DNN structure to optimize three objectives (accuracy, privacy, on-device resource costs). However, the problem is difficult due to the huge search space, the vague relation among different objectives (accuracy, privacy, on-device resource costs), and the retraining after compression. For example, a 11-layer AlexNet has 11 partition choices, with each convolutional layer having over 6 compression choices (\cite{howard2017mobilenets,sandler2018mobilenetv2,iandola2016squeezenet,han2015deep,li2016pruning} and no compression), and each fully-connected layer having at least 4 choices (\cite{lane2016deepx,bhattacharya2016sparsification,han2015deep} and no compression). The total number of configurations can be up to ten thousand. Therefore, an automatic feature selection framework is desired, as the manual DNN architecture design often fails to meet all user-defined criteria.

Our proposed framework automatically distributes the DNN structure across the mobile device and the cloud. %{\color{blue} Since image-based mobile applications such as healthcare \cite{white2014algorithms}, face recognition \cite{soyata2012cloud}, emotion recognition \cite{hossain2017emotion} has been widely adopted and its data security has raised serious concern, we focus on this type of applications in this paper. For each task,} 
We not only select which layer of features to be sent to the cloud but also explore the combination of compression techniques used to reshape the DNN on the device. Our method yields highly accurate models with optimized application performance while preserving input privacy. %The layer in selection and compression techniques constitute the new hyperparameters to be optimized, and 
Since the search space is huge and seeking an optimized solution is hard, we adopt a reinforcement learning (RL) based optimizer to search for the strategy to construct the intermediate-layer features to meet the objectives. 

More specifically, we search a distributed DNN structure by partitioning a base DNN and applying compression techniques to the DNN \textit{stub} residing at the device. The separated DNN parts are trained as an entirety to achieve a satisfactory level of accuracy and to prevent a generic adversary from inferring the private input information. Highlights of our contribution are as follows:
\begin{itemize}
	\item To the best of our knowledge, the work is among the first that preserves data privacy from the neural network structure perspective.
	\item We propose an RL-based optimizer which automatically searches for the best transformation and placement strategy of the DNN according to the customized performance criteria or system constraints.
	\item Our framework is evaluated on a variety of models and datasets. The experimental results show its superiority compared with the baseline in searching a distributed DNN structure meeting the requirements of accuracy, privacy, and on-device resource costs at the same time.
\end{itemize}

\section{Related work}
The related works fall into the following three parts.
\subsection{Running DNN with Mobile Devices}
While DNN and deep learning algorithms have been widely applied, it still faces significant computational challenges when migrated to thin mobile devices. A wide range of solutions has been proposed. From an {\em infrastructure} point of view, Lane \emph{et al.} \cite{lane2015can} were among the first to design a lower-power DNN inference engine on the device, taking advantage of both CPU and DSP chips to collaborate on mobile sensing and analysis tasks. Later in \cite{teerapittayanon2017distributed}, Teerapittayanon \emph{et al.} proposed to divide the DNN into different modules to deploy on the edge-cloud to improve the model accuracy and fault tolerance while keeping the communication cost low. Considering the resource constraints on the device, our work falls into the category of jointly deploying DNNs across mobile devices and the remote cloud to reduce resource cost.

Works such as \cite{padmanabha2018mitigating, han2016mcdnn, fang2018nestdnn, kang2017neurosurgeon, eshratifar2019jointdnn, laskaridis2020spinn} delved into the {\em trade-off} between deploying an accurate model and satisfying system performance constraints. By applying intelligent data grouping and task formulations, Iyer \emph{et al.} \cite{padmanabha2018mitigating} resolved the accuracy and latency trade-off in troubleshooting radio access networks using machine learning techniques. Han \emph{et al.} \cite{han2016mcdnn} systematically traded off DNN classification accuracy for resource use by adapting to high workloads with less accurate variants of models. Similarly, Fang \emph{et al.} \cite{fang2018nestdnn} dynamically selected the optimal resource-accuracy trade-off for each deep learning model to fit the system's available resources. Neurosurgeon \cite{kang2017neurosurgeon} proposed a lightweight scheduler to automatically partition DNNs to achieve low latency, low energy consumption, and high throughput. JointDNN \cite{eshratifar2019jointdnn} provided an energy-efficient method for on-device training and inference. Spinn \cite{laskaridis2020spinn} proposed a two-end progressive inference engine which can deliver higher performance over the state-of-the-art systems across diverse settings. Following the same principle, we could partially mitigate the resource-accuracy trade-off, but at the sacrifice of data privacy. Our work is among the first few that not only trains a neural network but also configure the deployment that fits best to our privacy objective.

\subsection{Privacy Threats in Deep Learning}
Abundant evidence has shown that the deployment of deep learning algorithms could potentially leak user privacy. Among them, we choose two typical privacy threats for investigation: feature inversion attack and property inference attack. The first category includes works such as \cite{mahendran2015understanding, dosovitskiy2016inverting}. Mahenran \emph{et al.} \cite{mahendran2015understanding} and Dosovitskiy \emph{et al.} \cite{dosovitskiy2016inverting} launched input reconstruction attacks on intermediate-layer features. Both works formulated the reconstruction as an optimization problem. Apart from intermediate-layer features, works like \cite{geiping2020inverting,zhu2019deep,zhao2020idlg} demonstrated the exchanged gradients during training can also recover participants’ training data.

%It has been demonstrated that a significant part of the input can be recovered even from features merely with high-level semantic information.

Examples of the second category include the following. Membership inference attacks \cite{shokri2017membership} could adversarially distinguish if a specific data record was used in training by exploiting the shadow models. Passive inference attacks \cite{ganju2018property} could infer unintended global properties of the training set merely from the trained models. In collaborative learning \cite{melis2019exploiting} and federated learning \cite{zhibo2019beyond,nasr2019comprehensive}, the adversary could infer properties of a subset of the training data in the process that multiple users jointly train a global model. Until recently, there is no effective defense against the above two attacks, and our work proposes no defense but rather seeks a neural network structure that meets users' privacy requirements against the two attacks.

\subsection{DNN Compression}
Compression is commonly used to reduce the complexity of DNN and the storage cost. Various compression techniques include weight compression \cite{lane2016deepx,bhattacharya2016sparsification,han2015deep}, convolution decomposition \cite{howard2017mobilenets,sandler2018mobilenetv2}, and special architecture layers \cite{iandola2016squeezenet}. For example, in \cite{han2015deep}, Han \emph{et al.} pruned unimportant model weights to compress the neural network. Along with quantization, their method has reduced the neural network size by $35$ times with almost no accuracy degradation. Unlike non-structured pruning in \cite{han2015deep}, structured pruning kept the layer-wise structures intact but shrank the size of the neural network by removing the entire layer or scaling down the kernel size or the filter number \cite{ashok2018n2n}. Targeted at a number of mobile platforms, Liu \emph{et al.} \cite{liu2018demand} trimmed down the network complexity with a number of compression techniques to satisfy resource constraints. While previous works mostly addressed the model accuracy and compression ratio, user privacy was rarely mentioned. Works such as \cite{zhao2018compress,carlini2017towards} investigate the combined effect of adversarial attacks and neural network compression which are remotely related to our work. %without defense approaches. \cite{carlini2017towards} even demonstrates defensive distillation does not significantly increase the robustness of neural networks.}

Our framework considers accuracy, privacy, and resource consumption at the same time from the perspective of neural network structures. In addition, we quantified the privacy leakage of our framework by conducting feature inversion attack and property inference attack.

\section{Preliminaries}
In this section, we will give a brief overview of the techniques adopted in this paper.

\subsection{Neural Architecture Search and AutoML}
Until now, the architectures of most neural networks are subject to manual selection, but there is an on-going trend to automatically search for optimal neural network structures, of which the accuracy could match the hand-designed one \cite{zoph2016neural,zoph2018learning,liu2018progressive}. Neural architecture search (NAS) \cite{zoph2018learning,liu2018progressive} aims to search ``cell" or ``block" which are stacked to construct models. To reduce the search space, \cite{ashok2018n2n} constructs networks by modifying a base model instead of building from scratch. Compressing a large neural network to a small one rather than growing one from scratch can shrink the search space. Inspired by this, we introduce the base DNN pool in our framework.

A number of search strategies can be used to explore the search space, such as random search, Bayesian optimization, evolutionary methods, reinforcement learning, and gradient-based methods \cite{elsken2019neural}. Among them, reinforcement learning has obtained competitive performance \cite{zoph2016neural}. Along this line, we adopt a reinforcement learning-based approach and carefully design its action space, state space, and reward function to effectively search for an optimal policy to find the desired DNN architecture.

\subsection{Reinforcement Learning}
Reinforcement learning is mostly concerned with how software agents take actions in an environment so as to maximize the cumulative reward. At each state, the agent takes action and communicates with the environment, which returns new observation and reward for the current state and action. The agent adapts its policy which returns a new action and the action would lead to a new state of the agent. The objective of the agent is to learn an optimal policy to maximize the reward accumulated in the long run.

Since the optimal policy is far too complicated to find, many methods are proposed to approximate one. For example, there are value-based methods \cite{mnih2015human}, policy-based methods\cite{williams1992simple} and combined methods like Actor-Critic\cite{lillicrap2015continuous}. In our work, we adopt Monte-Carlo policy gradient, a simple version of policy-based method, to search for the optimal DNN structure and placement strategy.

\section{Framework Overview}
\begin{figure}[h]
	\centering
	\includegraphics[width=0.7\linewidth]{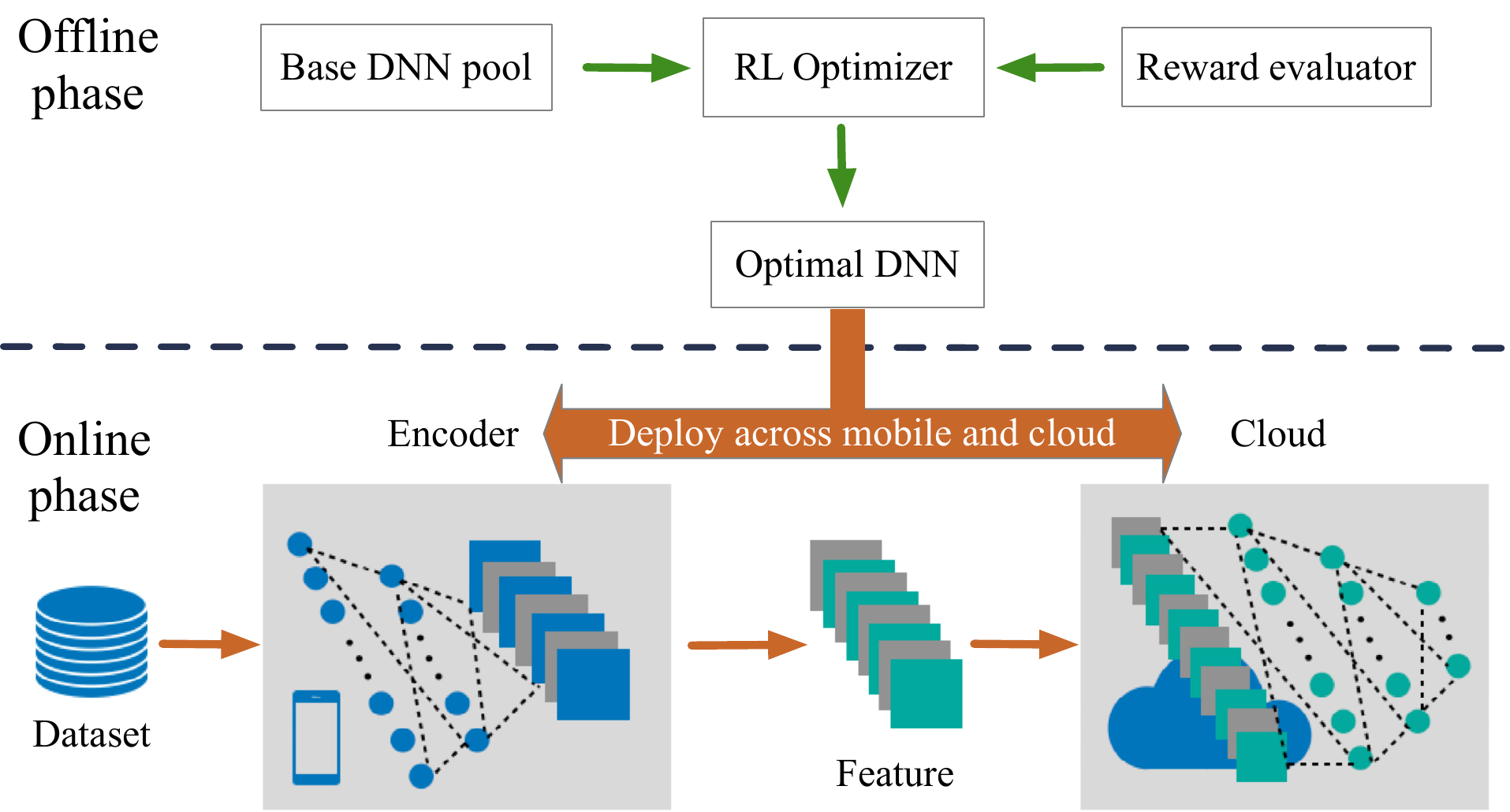}
	\caption{Framework overview.}
	\label{fig:framework}
	\centering
\end{figure}

Our goal is to seek the optimal partition and a combination of compression techniques that tailors a conventional DNN to one customized for the mobile-cloud framework, achieving high accuracy, privacy at the low cost of the device. Our framework consists of the following two phases shown in Fig.~\ref{fig:framework}.

\textbf{Offline phase.} The offline phase involves three modules: a base DNN pool, a reward evaluator, and a reinforcement learning based LSTM optimizer. The DNN pool contains most of the up-to-date standard deep neural networks, from which the users pick one based on targeting tasks. The base models in the DNN pool could be publicly pre-trained to reduce the re-training costs of the optimal DNN, but it is not a hard requirement. 
	
Given the base model structure as the input, the optimizer makes a decision on which layers to retain on the device and which layers to offload to the cloud, as well as the compression techniques applied per layer on the device. For presentation compactness, we refer to the DNN stub on the device as \textit{encoder}, and the remaining DNN part as \textit{cloud}. Each configuration of the model leads to a unique reward which takes accuracy, privacy, and resource costs into account and is evaluated by the reward evaluator. The weights of the optimizer get updated according to the reward value. The above three steps iterate until the optimizer converges to the optimal configuration, i.e., the optimal DNN. In a nutshell, our RL optimizer produces a configuration which 1) satisfies the mobile system requirement at runtime, 2) yields accurate inference results, and 3) preserves input privacy.

\textbf{Online phase.} In this phase, the optimal DNN is deployed across the mobile device and cloud for inference on private data. The user processes its data locally and sends the computed features to the cloud to complete the inference on its sensitive data.

To summarize, given a base DNN, we aim to search for an optimal strategy to {\em divide} and {\em compress} the DNN to obtain a DNN structure spanning across the mobile device and the cloud. Such a structure would meet our requirements on {\em model accuracy}, {\em data privacy}, and {\em system performance}, which we will illustrate in detail in the following section.

\section{Formulation}
\label{eqn:formulate}
As we have shown, in the offline phase, the optimizer is updated according to the reward incorporating accuracy, privacy, and resource costs.
%As we have shown earlier, accuracy, privacy, and resource utilization are seemingly conflicting goals to achieve within one neural network. Hence, we aim to search for an optimal DNN structure and placement to find a sweet spot in between these objectives. 
In this section, we will illustrate how we measure or evaluate each metric in our framework.

\subsection{Notation}
We use images as example studies but similar metrics can be applied to other cases. We define a dataset with $m$ training examples as $D=\{ (\mathbf{x}_1,y_1), \ldots, (\mathbf{x}_m, y_m) \}$, where $\mathbf{X} = \{\mathbf{x}_1,\ldots, \mathbf{x}_m\}$ denotes the set of input images and $\mathbf{y}=\{y_1, \ldots, y_m\}$ is the corresponding true label. The parameters of the encoder $e$ and the cloud $c$ are collectively represented as $\theta_e$ and $\theta_c$. The intermediate features output by $e$ is expressed as $\mathbf{M} = f_{e}(\mathbf{X}; \theta_e)$ whereas $c$ takes the features $\mathbf{M}$ as inputs, and outputs predictions $\hat{\mathbf{y}} = f_{c}(\mathbf{M}; \theta_c)$. 
\subsection{Accuracy}
The \textbf{accuracy} performance is evaluated both on $\theta_e$ and $\theta_c$ such that:
\begin{equation}
	\label{eqn:accuracy}
	\begin{aligned}
	A & = 1 - \frac{\sum_{i}1[\hat{y}_{i} \neq y_{i}]}{m}\\
	& = 1 - \frac{\sum_{i}1[f_{c}(f_{e}(\mathbf{x}_{i}; \theta_e); \theta_c) \neq y_{i}]}{m}.
	\end{aligned}
\end{equation}
Note that accuracy has nothing to do with where we divide between $e$ and $c$ as long as the two parts are trained jointly as an entirety.

\subsection{Privacy Losses}
\textbf{Threat model.} As there is no consensus on the measurement of \textit{privacy loss} of the neural network features in the current literature, we propose to evaluate privacy by how much a generic adversary could infer about $\mathbf{X}$ from the released features $\mathbf{M}$. The adversary constructs a neural network called {\em decoder }and trains the decoder on auxiliary dataset $\mathbf{X}^a$ and corresponding features $\mathbf{M}^a$ to evaluate privacy loss. In practice, we cannot control what the adversary obtains in the auxiliary dataset $\mathbf{X}^a$, and thus we assume a worst-case adversary who is able to train the decoder over $\mathbf{{X}^{a} = X}$. We introduce two types of threats as follows.

%We treat the decoder as an adversary who tries to infer private information from the released features $\mathbf{M}$. As there is no consensus on the measurement of \textbf{privacy loss} by the neural network features in the current literature, we propose to evaluate privacy by how much a generic adversary could infer about $\mathbf{X}$ beyond what contains in $\mathbf{y}$. We introduce two categories of attack to the intermediate-layer features by summarizing from the current literature.

\textbf{Feature inversion attack.} In this category of attack, the adversary tries to reconstruct the original input from features as in \cite{dosovitskiy2016inverting, zeiler2014visualizing, mahendran2015understanding}. For example, Dosovitskiy {\em et. al.} \cite{dosovitskiy2016inverting} uses an up-convolutional network to reconstruct images from features by minimizing the mean squared error (MSE) between the reconstruction images and original images. Similar to \cite{dosovitskiy2016inverting}, we adopt an up-convolutional network as the decoder with parameter $\theta_{d1}$, and train the decoder on auxiliary dataset $\mathbf{X}^a$ and evaluate the privacy loss. Such losses are gauged by the following metrics in this work.

The first metric is MSE. By this metric, the objective of the decoder is to minimize the MSE:
\begin{equation}
\label{eqn:ifr}
	\theta^{*}_{d1} = \arg \min_{\theta_{d1}} \sum_{\mathbf{x}_{i} \in \mathbf{X}^{a}} \|\mathbf{x}_{i} - f_{d}(f_{e}(\mathbf{x}_{i}; \theta_e); \theta_{d1}) \|^{2}_{2}.
\end{equation}
Given $\theta^{*}_{d1}$, the privacy loss is evaluated by MSE:
\begin{equation}
\label{eqn:p0}
P_0 = {1}/{\mathbb{E}[\| f_{d}(f_{e}(\mathbf{x}_{i}; \theta_e); \theta^{*}_{d1}) - \mathbf{x}_{i} \|^{2}_2]}.
\end{equation} 
The MSE gets lower when the reconstructed images are closer to the original ones, and thus the released features have a higher privacy loss. Besides the MSE, we also adopt the structural similarity index (SSIM) \cite{wang2004image} as the second metric to extract structural information from the recovered images. Different from MSE, a higher SSIM indicates a higher privacy loss. Hence, the objective of the decoder is to reconstruct the input by maximizing the SSIM value:
\begin{equation}
\label{eqn:fir}
\theta^{*}_{d1^{\prime}} = \arg \max_{\theta_{d1^{\prime}}} \sum_{\mathbf{x}_{i} \in \mathbf{X}^{a}} \textrm{SSIM}(\mathbf{x}_{i}, f_{d}(f_{e}(\mathbf{x}_{i}; \theta_e); \theta_{d1^{\prime}})).
\end{equation}
Likewise, we define the privacy loss as the similarity metric:
\begin{equation}
\label{eqn:p1}
P_1 = \mathbb{E}[\textrm{SSIM}(\mathbf{x}_{i}, f_{d}(f_{e}(\mathbf{x}_{i}; \theta_e); \theta^{*}_{d1^{\prime}}))].
\end{equation}

\textbf{Property inference attack.} In this category, the adversary infers from features about the input properties beyond what contains in the output, as in works such as \cite{shokri2017membership, ganju2018property, melis2019exploiting}. That indicates the adversary capability of deducting additional information irrelevant to the task. For example, let the task being the classification of a person's age ranges based on its facial image. We would not expect any party to learn any unrelated attribute, for example, whether the person wears glasses, from the released features. Hence the objective of the decoder is to train a classifier $\theta^{*}_{d2}$ to distinguish sensitive attributes $(x_{i}, y_{i}^{s})$ from features:
\begin{equation}
\label{eqn:ppi}
\theta^{*}_{d2} = \arg \min_{\theta_{d2}} \sum_{\mathbf{x}_{i} \in \mathbf{X}^{a}} 1[f_{d}(f_{e}(\mathbf{x}_{i}; \theta_e); \theta_{d2})\neq y_{i}^{s}].
\end{equation}
The privacy loss is defined as the inference accuracy on the unintended attributes:
\begin{equation}
\label{eqn:p2}
\begin{aligned}
P_2 = 1 - \sum_{\mathbf{x}_{i} \in \mathbf{X}^{a}}1[f_{d}(f_{e}(\mathbf{x}_{i}; \theta_e); \theta^{*}_{d2})\neq y_{i}^{s}] / \vert \mathbf{X}^{a}\vert,
\end{aligned}
\end{equation}
where $\vert \mathbf{X}^{a}\vert$ is the total number of instances in $\mathbf{X}^{a}$. Overall we try to mimic the real-world privacy threats to provide insights on how the neural network architecture design would affect the privacy performance. %In practice, we cannot control what the adversary obtains in the auxiliary dataset $\mathbf{X}^a$, and thus we assume a worst-case adversary, {\em i.e.}, $\mathbf{{X}^{a} = X}$, in training our privacy-preserving DNNs.

% system performance
\subsection{Resource Utilization}
Apart from accuracy and privacy, resource utilization is one major factor in configuring the encoder. The resource utilization is concerned with user experience, {\em i.e.,} how large space it takes (storage), how much energy it consumes (computation), and how fast the DNN processes the data (latency). We focus on storage, energy consumption, and latency at the device side. In the worst case, the base model is deployed as an entirety to the device which serves as the baseline. We aim to devise an encoder with low on-device resource cost considering the following factors.

\textbf{Storage} denotes the amount of on-device storage consumption:
\begin{equation}
\label{eqn:storage}
S_{1} = \max \{1 - \frac{\#\mathrm{params}(\mathrm{encoder})}{\#\mathrm{params}(\mathrm{base~model})}, 0\}
\end{equation}

\textbf{Computation} is represented by the total number of MACs (multiply–accumulate operation) of DNN operations \cite{liu2018demand}. Most MACs for inference lie in the convolutional (Conv) layers and the fully-connected (FC) layers, and are approximated as follows:.
\begin{equation}
\textrm{MAC}_{\textrm{conv}} = K \times K \times C_{in} \times C_{out} \times H_{out} \times W_{out},
\end{equation}
\begin{equation}
\textrm{MAC}_{\textrm{FC}} = C_{in} \times C_{out},
\end{equation}
where $K$ represents the kernel size, $H_{out}, W_{out}$ are the height and width of output feature, and $C_{in}, C_{out}$ denote the number of input and output channels of the layer. Therefore, computation is represented as:
\begin{equation}
\label{eqn:energy}
S_{2} = \max \{1 - \frac{\mathrm{MACs}(\mathrm{encoder})}{\mathrm{MACs}(\mathrm{base~model})}, 0\}
\end{equation}

\textbf{Latency} of running inference includes the running time on the device $\mathrm{T}_e$, the transfer latency from mobile to cloud $\mathrm{T}_t$ and the running time on the cloud $\mathrm{T}_c$. The cloud processing time is typically negligible compared with others and thus omitted. Formally, latency is formulated as:
\begin{equation}
\label{eqn:latency}
S_{3} = \max \{1 - \frac{\mathrm{T}_e + \mathrm{T}_t}{\mathrm{T}(\mathrm{total})}, 0\}.
\end{equation}
$\mathrm{T}(\mathrm{total})$ represents the running time if placing the entire model on the device. Since it is not practical to measure the real-world latency in our framework, we profile the latency beforehand with a similar approach to  \cite{wang2020context}.
%All these factors are relevant to the amount of computation on the device: a deeper or wider DNN stub takes larger storage space and most often involves more computation leading to longer inference time as well as higher energy consumption. For simplicity, we adopt the following two metrics to denote the resource utilization:

Essentially, with a larger $S_{1}$, $S_{2}$ or $S_{3}$, the storage, computation, or latency costs on the mobile device are lower, with more workload transferred to the cloud. %We certainly can use a more complex latency or energy model for better describing the performance gain, but that is out of the scope of this work. 

To sum up, {\em accuracy} $A$ is evaluated on the model jointly composed by the encoder and the DNN at the cloud. {\em Privacy loss} $P_{0}, P_{1}, P_{2}$ are defined as the amount of unintended information contained in the features released. The worse the adversary's performance in recovering inputs or inferring unintended properties of the input, the higher level of privacy the mechanism guarantees. {\em Resource utilization} $S_1, S_2$, $S_3$ are concerned with both the compression ratio and partitioning strategy. Among these metrics, privacy and accuracy are related to both model parameters and hyperparameters, while the resource utilization is mostly relevant to configuration hyperparameters such as the number of parameters per layer, layer type, input size, etc. In the following section, we will introduce our reward objective function composed of the metrics and the strategy design space.

\section{Reinforcement Learning Based Optimizer}
Our goal is to learn the optimal partition and compression strategy (policy) $\pi^{*}$ via reinforcement learning. The problem can be modeled as a Markov Decision Process (MDP) which is represented by a tuple $\mathcal{M} =  <\mathcal{S}, \mathcal{A}, \mathcal{T}, \gamma, r>$.

{\bf State:} $\mathcal{S}$ is the finite state space which contains all possible neural architectures deployed across the mobile devices and the cloud side. Let $s \in \mathcal{S}$ be a tuple list representing the architecture of a DNN. The length of $s$ equals the number of DNN layers. $s_i$ represents the $i$-th layer which is a tuple $<l,k,s,p,n>$. $l$ denotes the layer type, $k,s,p,n$ are the kernel size, stride size, padding length, and the number of output channel respectively.

{\bf Action:} A finite action set $\mathcal{A}$ transforms a state from one to another. We consider two types of actions in this work: the {\em partition} of the model across different devices and model {\em compression}. By compressing a base DNN to one fitting the mobile end, it not only shrinks down the search space but with some accuracy guarantee by the conclusion of \cite{han2015deep}. %{\color{red}For each DNN, we first select the partition and then select several compress actions. The number of compress actions is equal to the number of layers in the mobile devices which is determined by the partition action.}

{\bf Transition probability and discount factor:} Letting the transition probability space be $\mathcal{T}$, we take a deterministic transition since an action always transforms a DNN structure from one to another with probability $1$. And we set the discount factor $\gamma$ to $1$ to make each reward contribute equally to the final return.

{\bf Reward function:} The reward $r: \mathcal{S} \mapsto \mathbb{R}$ can be factorized as model accuracy, privacy, and resource utilization aforementioned. After taking each action, we obtain a new state --- a new DNN structure and its placement --- along with its associated reward $r$. %To explore more DNN structures at one run, we generate more than one new DNN within each episode and use their average reward as the reward for the action taken. 
Note that with each action taken, not all metrics would change. For instance, the action of partition does not change the model accuracy or the performance indicator, while compression would affect all metrics. Further, we define the intermediate reward to be zero, {\em i.e.}, the reward should only be given when the compression procedure is done for all layers, and any intermediate state would be rewarded $0$.

We consider a large reward should be given when the model achieves a high level of all three metrics. If all three metrics are treated equivalently, we can express the reward in the following form:
\begin{equation}
\label{eqn:reward}
\begin{split}
R &= R_{A} \times R_{P} \times R_{S} \\
&= \frac{A}{A_{\mathrm{base}}} \times (1 - P_{i}) \times S_{j}(2-S_{j}), ~~i \in \{0,1,2\}, j\in \{1,2,3\}.
\end{split}
\end{equation}
$R_{A}, R_{P}, R_{S}$ are all normalized between the value of $0$ and $1$. With $A_{\mathrm{base}}$ expressed as the accuracy of the base DNN, $R_{A}$ indicates the ratio between the accuracy of the optimal DNN and the baseline accuracy. Similarly, $R_{P}$ is the negative of privacy loss. The reward $R_{S}$ is expressed as a concave function of $S_j$ as we assume the performance gain grows drastically when $S_j$ is small but becomes marginal when $S_j$ is close to $1$. The assumption behind is valid as the performance gain usually grows marginal with the increase of resource utilization.

\subsection{LSTM-based Optimizer}
\begin{figure*}[!htbp]
	\centering
	\subfigure[Partitioning optimizer. $a_p$ is the cut action.]{
		\label{fig:cut_lstm}
		\includegraphics[width=0.48\textwidth]{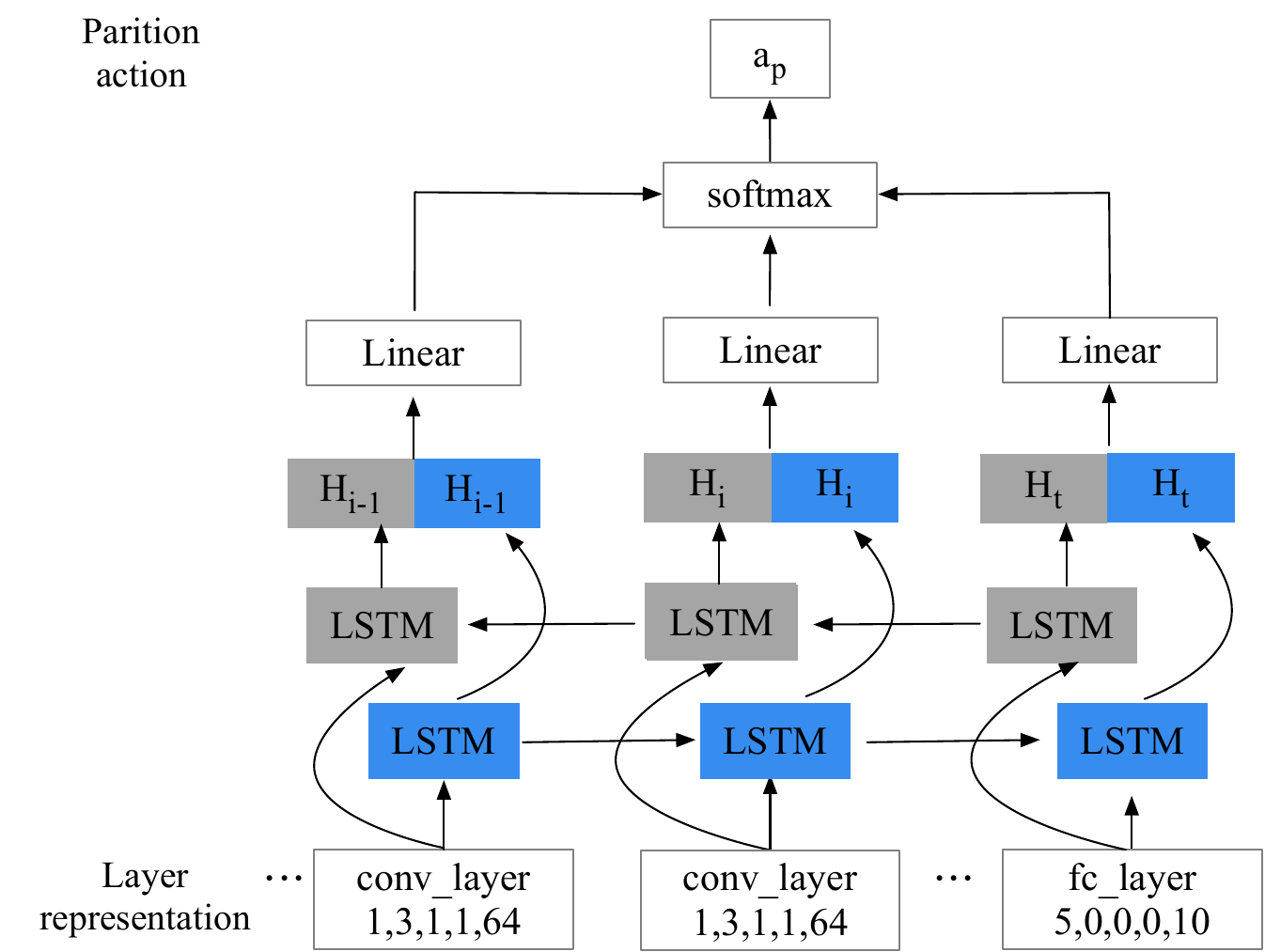}
	}
	\subfigure[Compression optimizer. $a_{c;i}$ is the compression action for layer $i$.]{
		\label{fig:com_lstm}
		\includegraphics[width=0.48\textwidth]{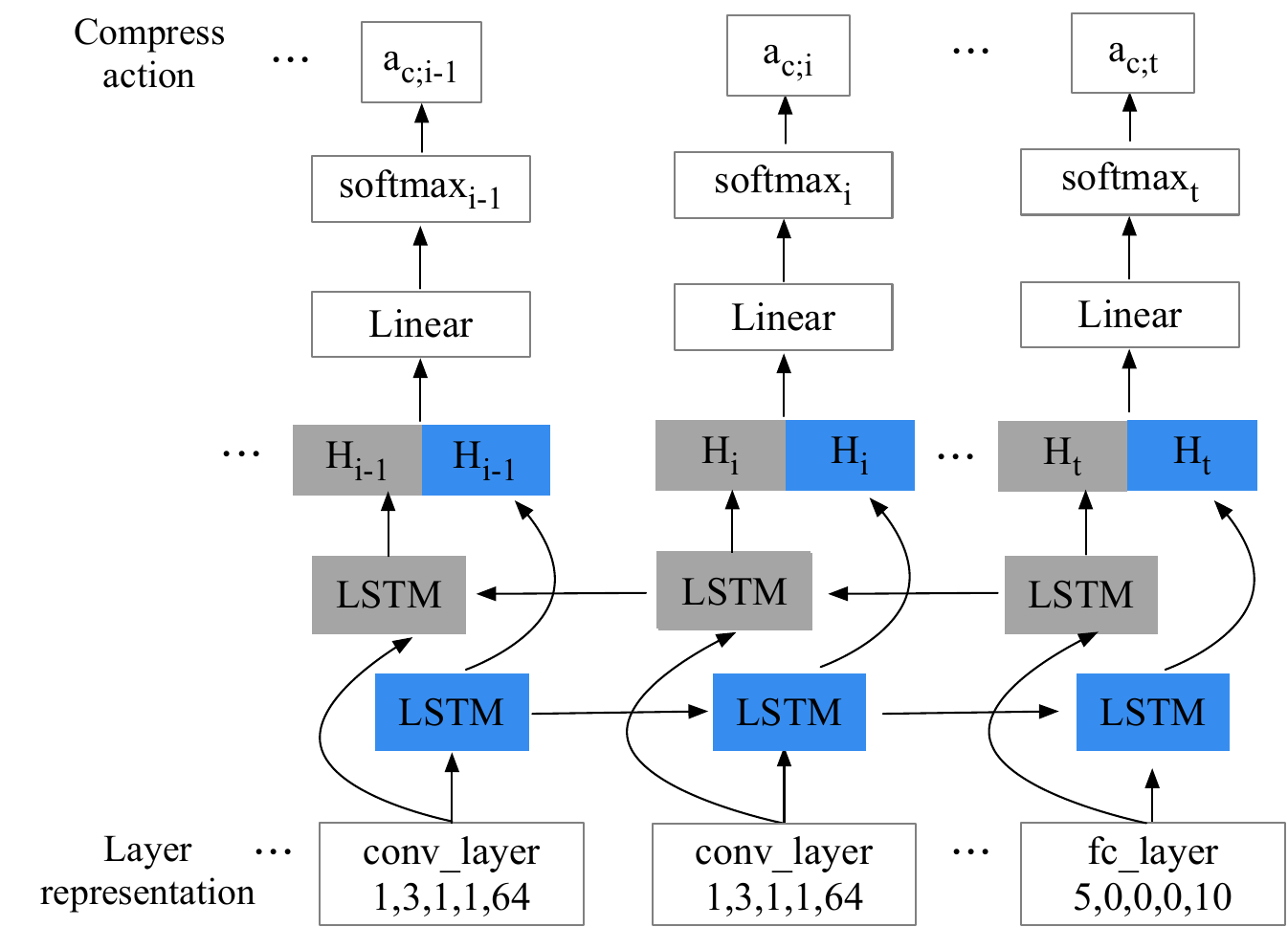}
	}
	\caption{Reinforcement learning based optimizer. Inputs are the layer representation and $t$ is the total number of DNN layers. Outputs are actions. $H_i$ represents the hidden states.}
	\label{fig:lstm}
\end{figure*}

Recurrent neural networks have been widely applied to neural architecture search \cite{chu2019fast,ashok2018n2n,zoph2016neural}, for their capability to process hyperparameters expressed as strings. We adopt the bidirectional LSTM, a type of recurrent neural network, as the decision engine in our reinforcement learning optimizer. 

We implement two LSTM-based optimizers --- a partition and a compression optimizer, the structure of which are given in Fig.~\ref{fig:lstm}. Each LSTM layer takes the current representation of the DNN layer as input, feeds them into a forward LSTM and a backward LSTM respectively to compute the corresponding hidden states $H_{i}$. As the change of one layer can affect the layer prior to it or the layer following it, the design of the backward and forward LSTM ensures such influence is reflected by the network. Then the hidden states are fed into linear layers. For the partition LSTM optimizer, the linear layer outputs are fed to the softmax layer to produce the partition decision $a_p$ with the highest probability. For the compression LSTM optimizer, the output size of the linear layer is equal to the number of all possible compression techniques which includes the case when no compression is applied. Specifically, a softmax layer follows the linear layer and produces the compression action $a_{c;i}$ with the highest probability for the $i$-th DNN layer. Given $a_{c;i}$ for all DNN layers and $a_p$, we are able to transform one DNN model to a new one, and its reward is updated correspondingly. Note that the compression techniques are only applied to the encoder since the cloud side has abundant computation power.

\subsection{Adversarial Retraining}
With each set of hyperparameters (and the DNN placement) as the state, we compose a neural network and retrain it. According to Eq.~\ref{eqn:reward}, the reward consists of $R_S, R_A$ and $R_P$. While $R_S$ can be obtained before retraining, the latter two can only be evaluated after retraining. As for calculating the accuracy and privacy, we apply two retraining strategies w.r.t. different threat models. 

\textbf{Reactive adversary.} We first consider a reactive adversary who manipulates the system by intentionally sending inputs to the encoder and retrieving the released intermediate-layer features. To mimic such an adversary, we train a decoder over the set of input-feature pairs and record the best performance achieved by the decoder in terms of the input reconstruction error or the property inference error ( Eq.\ref{eqn:p0},~\ref{eqn:p1} and \ref{eqn:p2}). The errors are adopted to gauge the privacy level that the encoder can achieve.

\textbf{Proactive adversary.} Also, we assume a proactive attack in which the adversary can interact with the training of the composed DNN. Particularly, the adversary trains a decoder against the current encoder to minimize the encoder's privacy gain. Our encoder is aware of such an adversary and thus it is trained to maximize its privacy gain and final output accuracy. In a nutshell, the encoder and decoder pit against each other to achieve a saddle point in the following minimax problem: 
\begin{equation}
	\min_{\theta_{e}, \theta_{c}} \max_{\theta_{d}} (- A + P_{i}), ~i \in \{0,1,2\}.
\end{equation}
In the equation above, the model accuracy $A$ is associated with both $\theta_{e}$ and $\theta_{c}$ whereas the privacy loss $P_{i}$ is a function of $\theta_{e}$ and $\theta_{d}$. The formulation resembles the training of a generative adversarial network (GAN) where the encoder and decoder play the roles of discriminator and generator respectively. In practice, it is hard to train such a minimax loss as the decoder usually converges much faster than the encoder and the cloud. Hence, we tune hyperparameters to facilitate the convergence of the minimax problem.

Overall, in the reactive adversary model, as the user and the adversary are not aware of the encoding or decoding strategies of the other, both of them are weak. As to the proactive adversary model, both the user and the adversary are stronger: the adversary, being aware of the power of the user, trains a decoder to compromise the user privacy, whereas the user seeks satisfactory privacy performance against such a worst-case adversary. The adversarial retraining process is shown in Alg.~\ref{alg:adv_training} where $N_p, N_{cd}, N_d$ are the number of training epochs.
\begin{algorithm}
	\renewcommand{\algorithmicrequire}{\textbf{Input:}}
	\renewcommand{\algorithmicensure}{\textbf{Output: }}
	\caption{Adversary Retraining $(e, c, d)$}
	\begin{algorithmic}[1]
		\REQUIRE Encoder $e$, cloud $c$, decoder $d$.
		\ENSURE Accuracy $A$, privacy loss $P$.
		\IF{reactive adversary}
		\STATE Train $e$ and $c$ for $N_{ec}$ epochs for $\max \limits_{\theta_{e},\theta_{c}} A$.
		\STATE Train $d$ for $N_d$ epochs for $\max \limits_{\theta_{d}}P$.
		\ENDIF
		\IF{proactive adversary}
		\FOR {$i = 1$ to $N_p$ epochs}
		\STATE Train $d$ for $N_d$ epochs for $\max \limits_{\theta_{d}}P$.
		\STATE Train $e$  and $c$ for $N_{ec}$ epochs for $\min \limits_{\theta_{e},\theta_{c}} (-A+P)$.
		\ENDFOR
		\ENDIF
		\RETURN Accuracy $A$, privacy loss $P$.
	\end{algorithmic}
	\label{alg:adv_training}
\end{algorithm}

\subsection{Optimization}
REINFORCE policy gradient algorithm \cite{williams1992simple} is applied to update the LSTM-based optimizer to find the optimal partition policy $\pi_{p}$ with parameters $w_p$ and compression policy $\pi_{c}$ with parameters $w_c$ during offline phase. Since the updating processes are the same for the two policies, we use $\pi, w$ to denote the policy and parameters without causing confusion. The objective function for the policy network is the expected reward over all sequences of actions $a_{1:L}$, {\em i.e.}:
\begin{equation}
\label{eq:objective}
J(w)=E_{a_{1:L}\sim \Pi(w)}[R].
\end{equation}
For each episode, $\pi$ is updated as follows:
\begin{equation}
\label{eqn:grad}
\nabla_{w}J(w) \approx \frac{1}{M}\sum_{j=1}^{M}\sum_{t=1}^{L}[\nabla_{w}\log \Pi(w)(a_t|h_t)R_j],
\end{equation}
where $L$ is the length of the trajectory. For partition policy, $L=1$ as the DNN model is partitioned at only one spot. For compression policy, $L$ is the number of encoder layers since we only apply compression techniques to the DNN stub on device. $M$ is the number of training samples per episode and we use the average reward to update the policy. $\Pi(w)(a_t|h_t)$ is the probability of selecting action $a_t$ at given hidden state $h_t$, and $R_j$ is the reward of the $j^{th}$ sample. We adopt a state-independent baseline function to reduce the variance in training:
\begin{equation}\label{eq:addBias}
\nabla_{w}J(w) \approx \frac{1}{M}\sum_{j=1}^{M}\sum_{t=1}^{L}[\nabla_{w}\log \Pi(w)(a_t|h_t)(R_j-b)],
\end{equation}
where the baseline $b$ is the exponential moving average of the previous rewards.

Alg.~\ref{alg:rla} summarizes the reinforcement learning optimization procedure. $N$ is the number of episodes. Line 4-9 get the partition and compression actions. Line 10-12 generate new DNN model and calculate the reward. Line 15-16 update the policy network.

\begin{algorithm}
	\renewcommand{\algorithmicrequire}{\textbf{Input:}}
	\renewcommand{\algorithmicensure}{\textbf{Output: }}
	\caption{Reinforcement Learning $(\mathcal{S}, \mathcal{A}, \mathcal{T}, r, \gamma)$}
	\begin{algorithmic}[1]
		\REQUIRE a base DNN structure
		\ENSURE a new DNN configuration
		\STATE Get the initial state $s_0$ of the base DNN
		\FOR {$i = 1$ to $N$}
		\FOR {$j = 1$ to $M$}
		\STATE The initial state for the $j^{th}$ sample: $s_{j, 0} = s_{0}$;
		\STATE Get partition action: $a_{p;j} \sim \pi_{p}(s_{j, 0}; w_{p;i-1})$
		\FOR {$t = 1$ to $L$}
		\STATE Get compress action: $a_{c;j,t} \sim \pi_{c}(s_{j,t-1}; w_{c;i-1})$;
		\STATE $s_{j,t} \leftarrow \mathcal{T}_{c}(s_{j,t-1},a_{c;j,t})$;
		\ENDFOR
		\STATE Generate the new DNN configuration according to $a_{p;j}$ and $a_{c;j,1},\cdots,a_{c;j,L}$.
		\STATE Calculate $A$, $P$ (call Alg.~\ref{alg:adv_training}) and $S$ (Eq.~\ref{eqn:storage},~\ref{eqn:energy},~\ref{eqn:latency})
		\STATE Calculate reward $R_j$ (Eq.~\ref{eqn:reward})
		\ENDFOR
		\STATE $R \leftarrow \frac{1}{M} \sum_{j=1}^{M} R_j$;
		\STATE Update partition policy: $w_{p;i} \leftarrow \nabla_{w_{p;i-1}}J(w_{p;i-1})$;
		\STATE Update compression policy: $w_{c;i} \leftarrow \nabla_{w_{c;i-1}}J(w_{c;i-1})$;
		\ENDFOR
	\end{algorithmic}
	\label{alg:rla}
\end{algorithm}

\begin{table*}[!htbp]
	\centering
	\caption{Experiment Setup}
	\scriptsize
	\begin{tabular}{ c | c | c | c | c | c }
		\toprule
		Dataset & Base model & Classification task & Attack type & Reactive & Proactive \\ \hline
		CIFAR-10 & \begin{tabular}[c]{@{}c@{}}VGG11, VGG13, VGG16, AlexNet,\\  Lenet, MobileNet v2, ResNet50, \\ Inception v3, Wide Resnet28-10\end{tabular} & 10 classes & Feature inversion  & $\surd$ & $\surd$ \\ \hline
		Tiny ImageNet & Resnet18 & 200 classes & Feature inversion & $\surd$ & \\ \hline
		CIFAR-100 & VGG11, VGG13, VGG16 & 20 superclasses & Property inference & & $\surd$ \\ \hline
		Purchase100 & FCN & 100 classes & Property inference & $\surd$ & \\ 
		\bottomrule
	\end{tabular}
	\label{tab:setting}
	\centering
\end{table*}

\begin{table*}[!htbp]
	\centering
	\caption{Compression Techniques}
	\scriptsize
	\begin{tabular}{c | c | c | c  }
		\toprule
		Name & Replaced Structure  & New Structure & Applied Layer Types \\ \hline
		$F_{1}$~(SVD)\cite{lane2016deepx} & $m \times n$ weight matrix & $m \times k$ and $k \times n (k \ll m)$ weight matrices  & FC layer  \\ \hline
		$F_{2}$~(KSVD)\cite{bhattacharya2016sparsification} & same above & same above with sparse matrices & FC layer  \\ \hline
		$C_{1}$~(MobileNet)\cite{howard2017mobilenets} & Conv layer & $3 \times 3$ depth-wise Conv layer and $1 \times 1$ point-wise Conv layer & Conv layer \\ \hline
		$C_{2}$~(MobileNetV2)\cite{sandler2018mobilenetv2} & Conv layer & same above with additional point-wise Conv layer and residual links & Conv layer  \\ \hline
		$C_{3}$~(SqueezeNet)\cite{iandola2016squeezenet} & Conv layer & a Fire layer & Conv layer  \\ \hline
		$W_{1}$~(Pruning)\cite{han2015deep} & any layer & insignificant parameters pruned layer & FC or Conv layer  \\ \hline
		$W_{2}$~(Filter Pruning)\cite{li2016pruning} & Conv layer & insignificant filters pruned Conv layer &  Conv layer \\
		\bottomrule
	\end{tabular}
	\label{tab:com}
\end{table*}

\begin{figure}[!htp]
	\centering
	\includegraphics[width=0.7\linewidth]{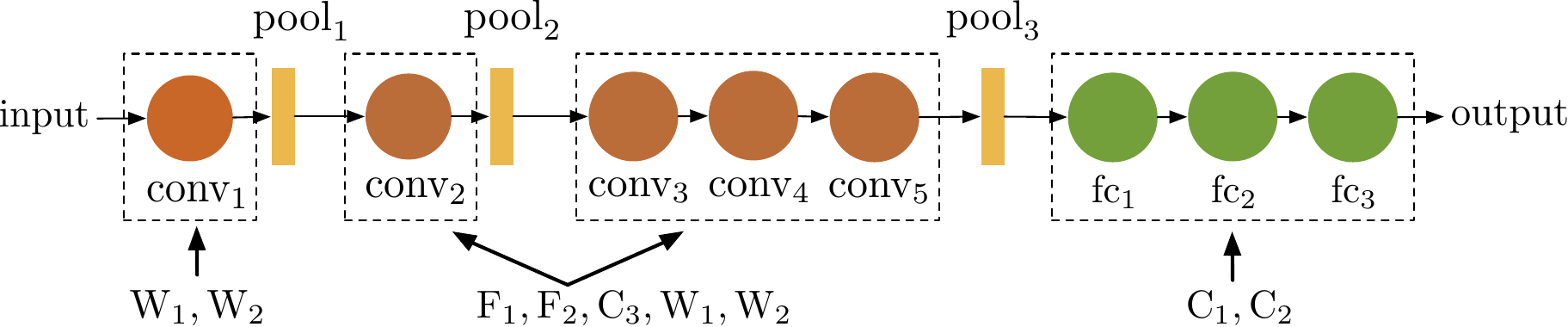}
	\caption{An illustration of how different layer compression techniques are applied to AlexNet.}
	\label{fig:alexnet_compress}
	\centering
\end{figure}

\section{Evaluation}
In this section, we test the performance of our RL-based optimizer in terms of accuracy, privacy and, resource utilization. The experiments are conducted on a number of datasets and classic neural networks.

\subsection{Setup}
The evaluation setup is provided in Tab.~\ref{tab:setting}. We test our RL-based optimizer against two types of attacks: feature inversion and property inference attack. The task on CIFAR-10 and ImageNet is image classification on sensitive images. The goal is to prevent the adversary from inverting the original input through the feature. For any DNN selected, we evaluate its accuracy by the classification accuracy, and the privacy by the SSIM \cite{wang2004image} of the reconstructed inputs. The task on CIFAR-100 is to classify the images into 20 superclasses with the finer classes as the private features.  We evaluate the model accuracy by the classification accuracy on superclasses, and the privacy by the classification error rate of the fine-class attributes. Purchase100 is a tabular dataset containing 600 binary attributes with 100 labels. The objective is to prevent input attributes reconstruction while learning a model with high classification accuracy.

\subsection{Implementation Details}
Our RL-based optimizer is implemented by the machine learning framework {\tt PyTorch} 1.5.0 and all experiments are done on Intel Xeon Processor with GPU GeForce RTX 2080 Ti.

{\bf Compression techniques.} We implement the most common compression techniques listed in Tab.~\ref{tab:com}. For better understanding, Fig.~\ref{fig:alexnet_compress} shows how different compression techniques can be applied to AlexNet as an example.

\textbf{Hyper-parameters.} We tried several combinations of hyper-parameters and choose the set leading to the best results. The default hyper-parameters for Alg.~\ref{alg:adv_training} are $N_{ec}=10$, $N_p=5$, and $N_d=10$.  We use SGD optimizer to re-train the new DNN model and the learning rate is 0.001. For RL optimizer training, the learning rate of partition and compression policy is selected as 0.03 and 0.003. The rollout number $M$ is set to 1, 5, 10. The optimizer of the policy network is Adam and we train the policy network for 200 episodes.

{\bf Decoder Structure:} For the feature inversion attack, we use an up-convolutional network similar to \cite{dosovitskiy2016inverting} to reconstruct images from features. The decoder adopts an inverse structure of the encoder. For example, the inverse of convolutional layers is transposed convolutional layers and pooling layers correspond to up-sampling layers. For the property inference attack, the decoder is a simple classifier to classify the unintended attributes.

{\bf Speedup:} We apply several tricks to reduce the training time. First, after getting partition and compression actions, we assign hash code to each generated model and store the corresponding reward in a memory pool. In this way, we can avoid redundant computation and significantly reduce training time. Second, in our experiments, the base models are pre-trained which requires less retraining time. Last, we adopt transferability across different models and datasets to avoid training policy network from scratch each time.

\subsection{Results}
We first verify the impact of different partition and compression strategies on accuracy and privacy. Second, we present results achieved by our RL-based optimizers along with baseline comparison to show the effectiveness of our method. Then, we discuss the transferability of our framework, including transferring policy trained on small DNN to large ones, and transferring policy network trained on small datasets to larger ones. The possibility of transfer permits less offline searching time for larger search space. Finally we deploy our algorithm on Nvidia TX2 to demonstrate its real-world performance. For presentation compactness, the combination of compression techniques applied is in the form of $applied\ layer : compression \ technique$.

\subsubsection{Accuracy \& Privacy vs. Compression \& Partition}
As aforementioned, different choices of configuration would incur different tradeoffs between accuracy and privacy. We support this claim with experiments on a variety of fixed configurations.
	
\textbf{Accuracy vs. Compression.}
{\em Compression Ratio} is defined as
$$
\textrm{CR} = 1 - \frac{\#\mathrm{params}(\textrm{compressed~DNN})}{\#\mathrm{params}(\textrm{base~DNN})}
$$
to indicate how much the model is compressed. We take AlexNet for example: when we divide AlexNet at the 12th layer and apply different combinations of compression techniques up to the 12-th layer, we obtain different CRs and accuracies as shown in Tab.~\ref{tab:compression}. We observe that different compression techniques lead to varied accuracies. C1 brings only 1.62\% reduction while the combination of W2 and C1 results in an unacceptable 6.06\% degradation. In general, higher CRs leads to lower resource cost but lower accuracy.
\begin{table}[!htbp]
	\centering
	\scriptsize
	\caption{Different compression techniques lead to different compression ratios and accuracies on AlexNet, CIFAR-10.}
	\begin{tabular}{c | c  c  c  c}
		\toprule
		$A_{\textrm{base}}$ & Compression Techniques & CR & $A$ & $\triangle A$ \\ \hline
		\multirow{5}*{0.8679}
		& 10:C1 & 0.0224 & 0.8538 & -1.62\% \\ \cline{2-5}
		& 0:W1 3:W1 6:W1 8:W1 10:W1 & 0.0645 & 0.8393 & -3.30\%\\ \cline{2-5}
		& 0:W1 3:W1 6:W1 8:W1 10:C1 & 0.0701 & 0.8251 & -4.93\%\\ \cline{2-5}
		& 0:W2 3:W2 6:W2 8:W2 10:C1 & 0.0743 & 0.8153 & -6.06\%\\ \cline{2-5}
		& 3:C3 6:C3 8:C3 10:C3 & 0.0859 & 0.8291 & -4.47\% \\ \hline
	\end{tabular}
	\label{tab:compression}
\end{table}
\begin{figure*}[!htbp]
	\centering
	\subfigure{
		\includegraphics[width=0.48\linewidth]{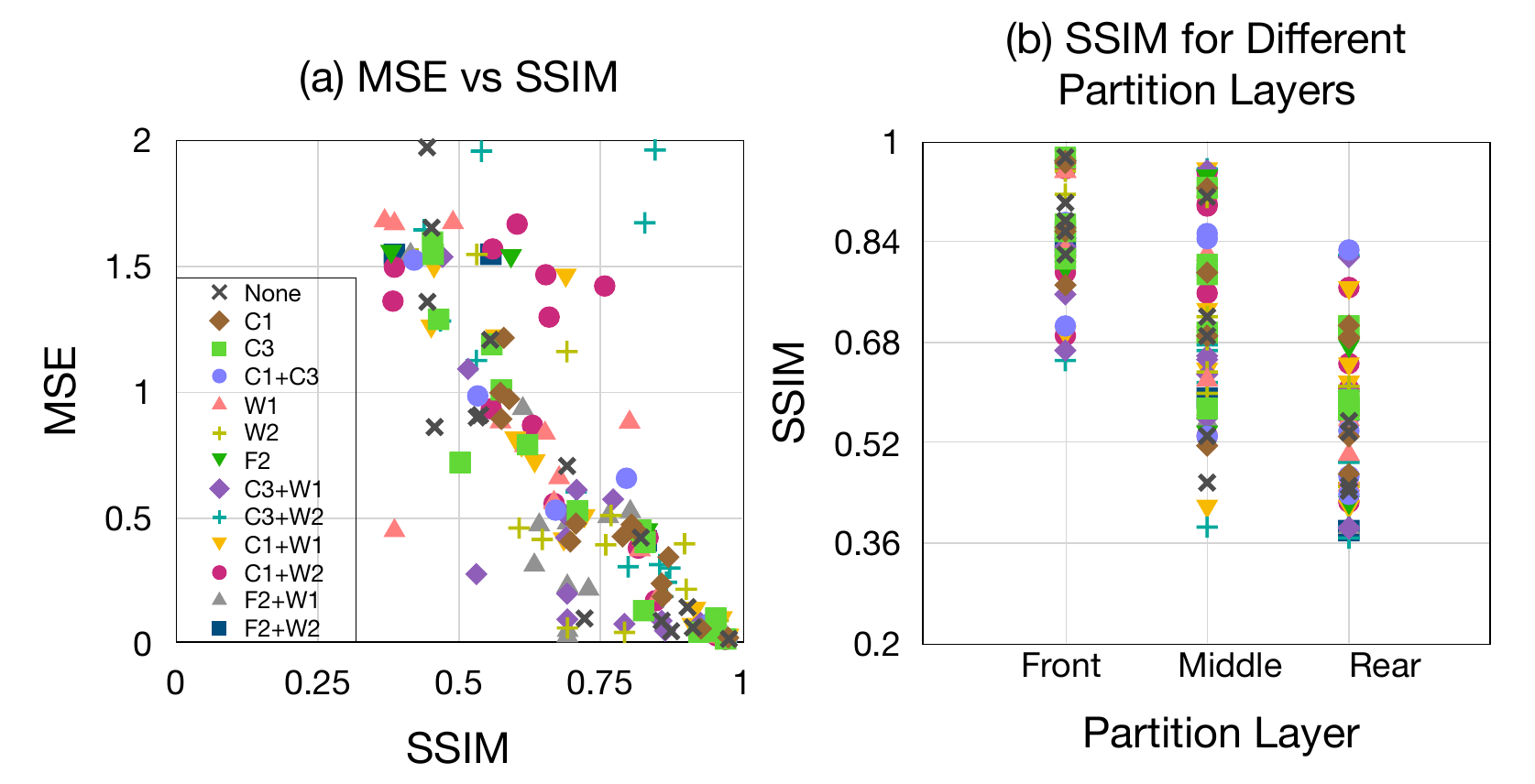}
	}
	\subfigure{
		\includegraphics[width=0.48\linewidth]{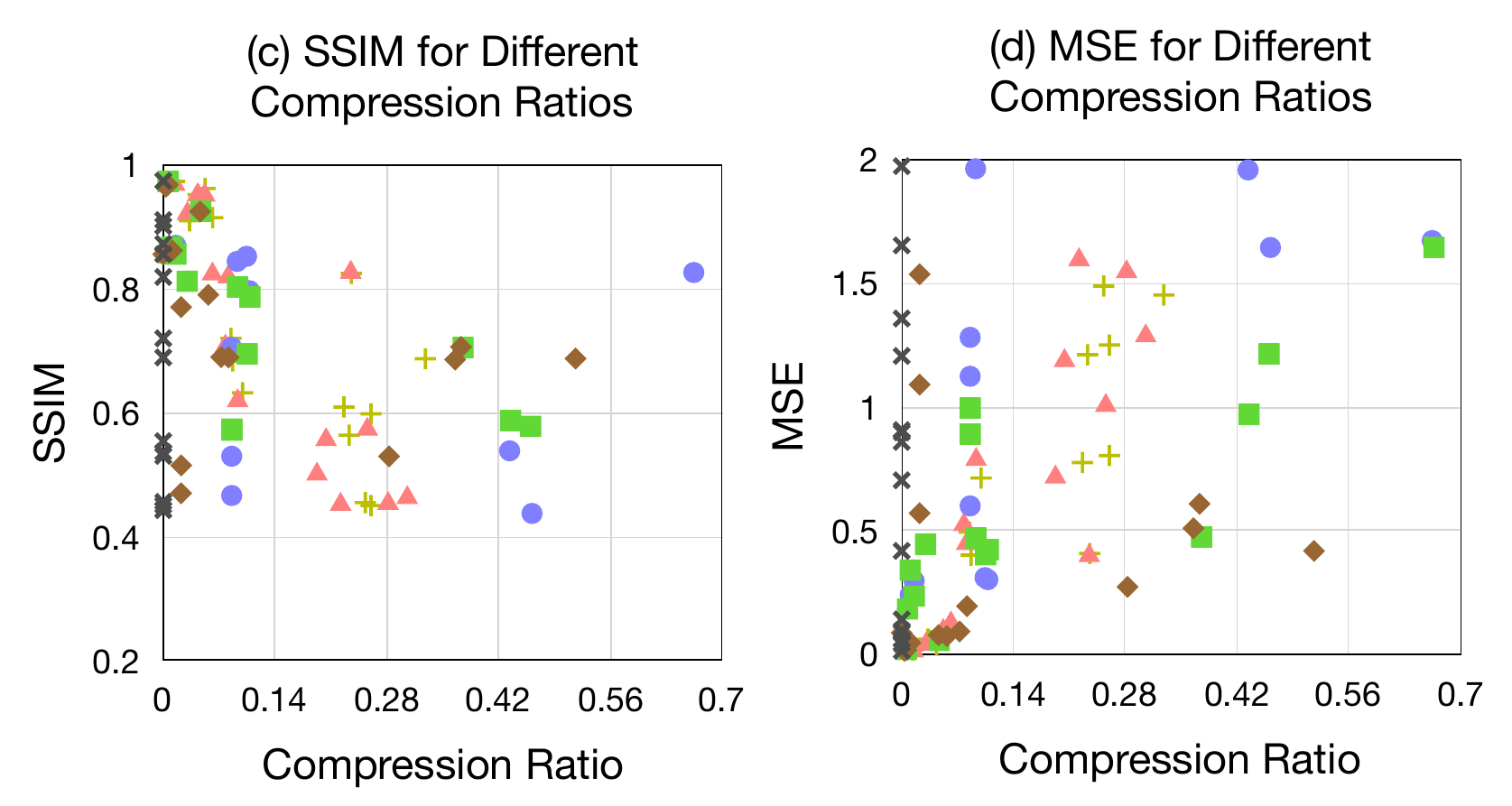}
	}
	\caption{
		The impact of compression and partition on privacy measured by SSIM and MSE on CIFAR-10. (a) (b)(c)(d) share the same legend. (a): SSIM and MSE are negatively correlated. (b): SSIM is lower when the partition layer is closer to the output layer of the model. (c): SSIM is inversely proportional to the compression ratio. (d): MSE is proportional to the compression ratio.}
	\label{fig:privacy_cr_cut}
\end{figure*}
\begin{figure*}[!htbp]
	\centering
	\includegraphics[width=0.8\linewidth]{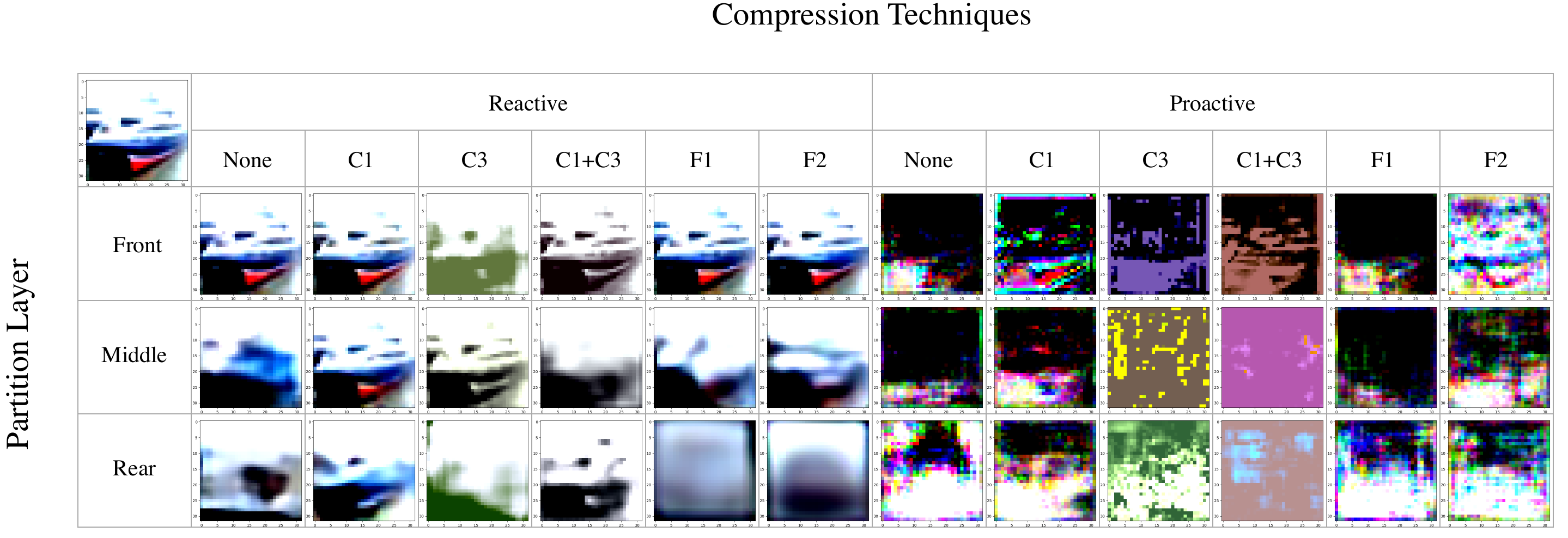}
	\caption{The visualization effect of reconstructed inputs reversed from different partitioning layers under a variety of compression techniques on VGG-13, CIFAR-10. %Reactive means the decoder only passively reconstructs the input from features. Proactive means that the decoder and encoder pit against each other for retraining the model. We observe that the reconstruction effect is poor when the partitioning layer is close to the output layer, especially when the model is compressed. Proactive decoder has a worse reconstruction effect as the encoder is enhanced with retraining.
	}
	\label{fig:reconst}
\end{figure*}
\textbf{Privacy vs. Compression and Partition.} We implement reactive adversary on VGG-11, VGG-13, VGG-16, LeNet, AlexNet, and ResNet-18 to show how partition and compression ratios affect the privacy. We select three representative positions to partition the neural network: front (closer to the input layer), middle, and rear (closer to the output). Fig.~\ref{fig:privacy_cr_cut}-(a) shows the consistency of the two metrics, {\em i.e.,} a higher SSIM usually indicates a lower MSE and thus a lower privacy level. In Fig.~\ref{fig:privacy_cr_cut}-(b), features closer to the rear have a lower SSIM and thus a higher privacy level. The observation concurs with the intuition that rear features contain less input information and is harder to attack. It is also observed in Fig.~\ref{fig:privacy_cr_cut}-(c)(d) that the more we compress the neural network, the harder it is for the decoder to reconstruct inputs from the features, mostly due to less expressive neural network features.

We also display the visualization effect of each reconstructed input restored from different features in Fig.~\ref{fig:reconst}. From left to right, each image is reconstructed from features compressed by different techniques, when reactive or proactive training is applied. From top to bottom, each image is reconstructed from different partitioning layers. Overall, features at front have a better reconstruction effect, whereas it is hard to reconstruct input from proactively trained features. The latter shows in proactive training, the encoder gradually learns how to shield information from the decoder.

Above all, different compression techniques and partitioning strategies indeed have various impacts on the accuracy and privacy. We observe that accuracy, privacy, and resource utilization are seemingly conflicting goals within one neural network from these findings.

\subsubsection{RL-based Optimizer}
We will show the results of our RL-based optimizer and the comparison with three baselines.

\begin{figure*}[!htbp]
	\centering  %图片全局居中
	\subfigure[VGG11 Results on CIFAR-10]{
		\label{fig:rl_1}
		\includegraphics[width=0.32\textwidth]{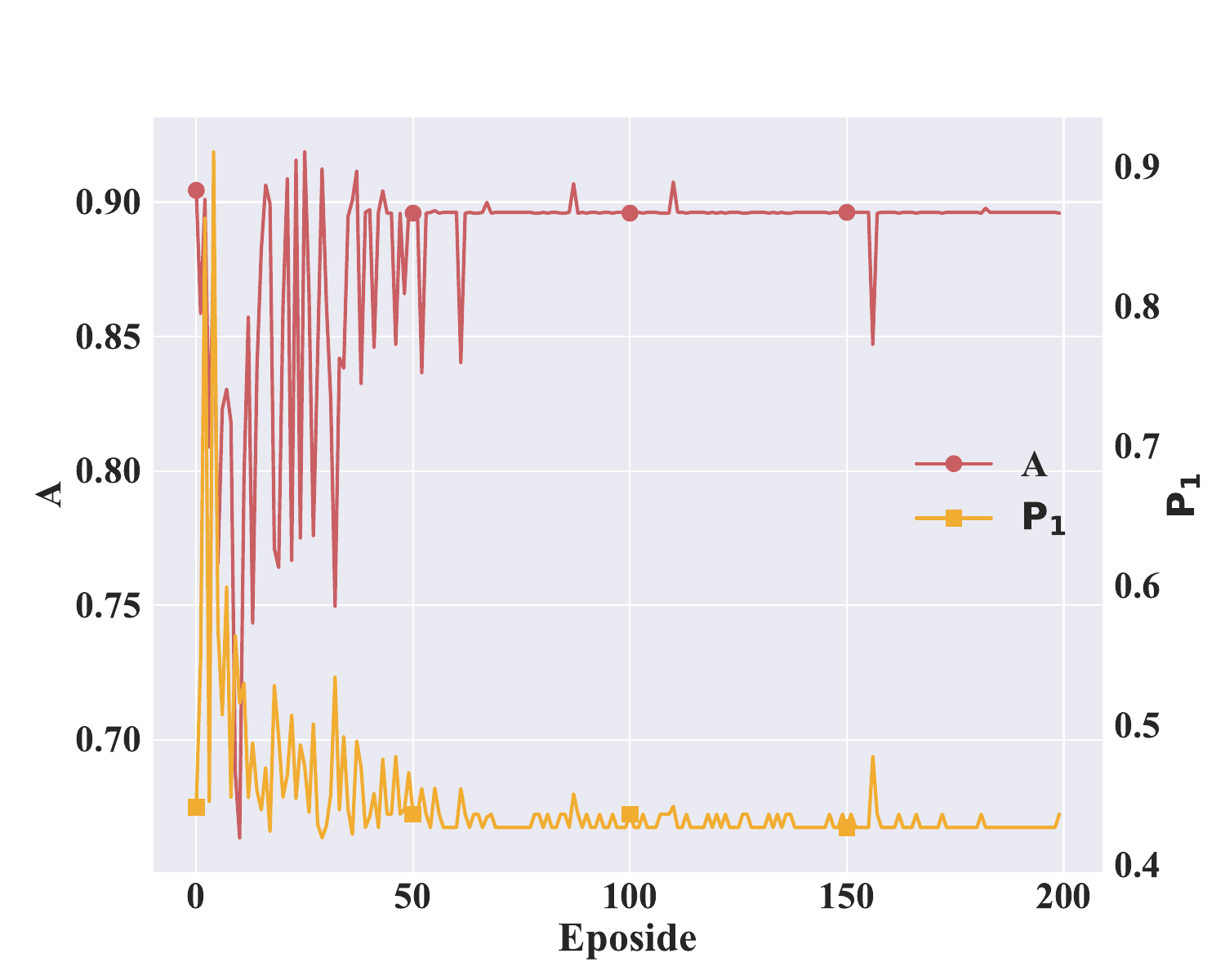}}
	\subfigure[VGG11 Results on CIFAR-10]{
		\label{fig:rl_2}
		\includegraphics[width=0.32\textwidth]{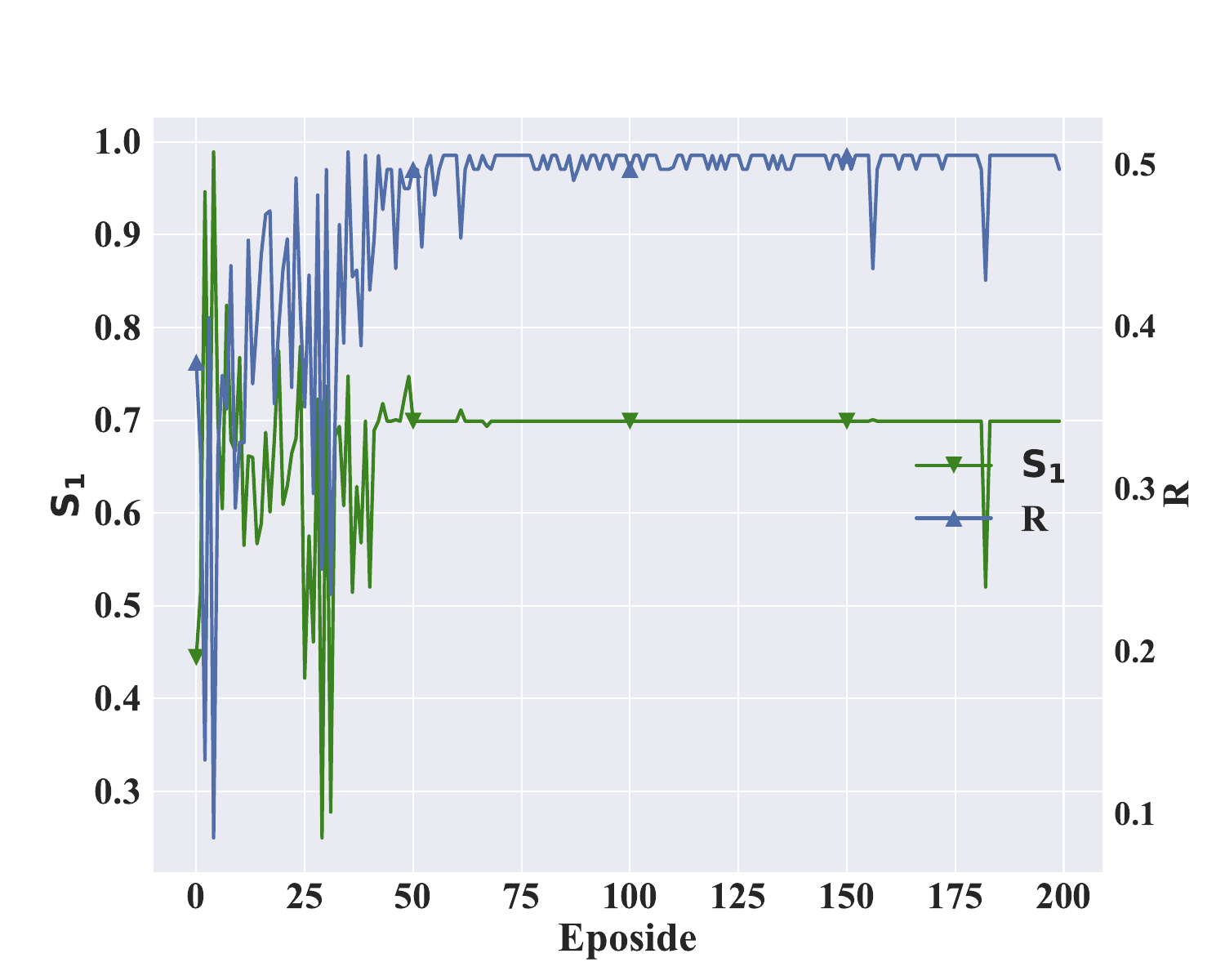}}
	\subfigure[VGG16 Results on CIFAR-10]{
		\label{fig:rl_3}
		\includegraphics[width=0.32\textwidth]{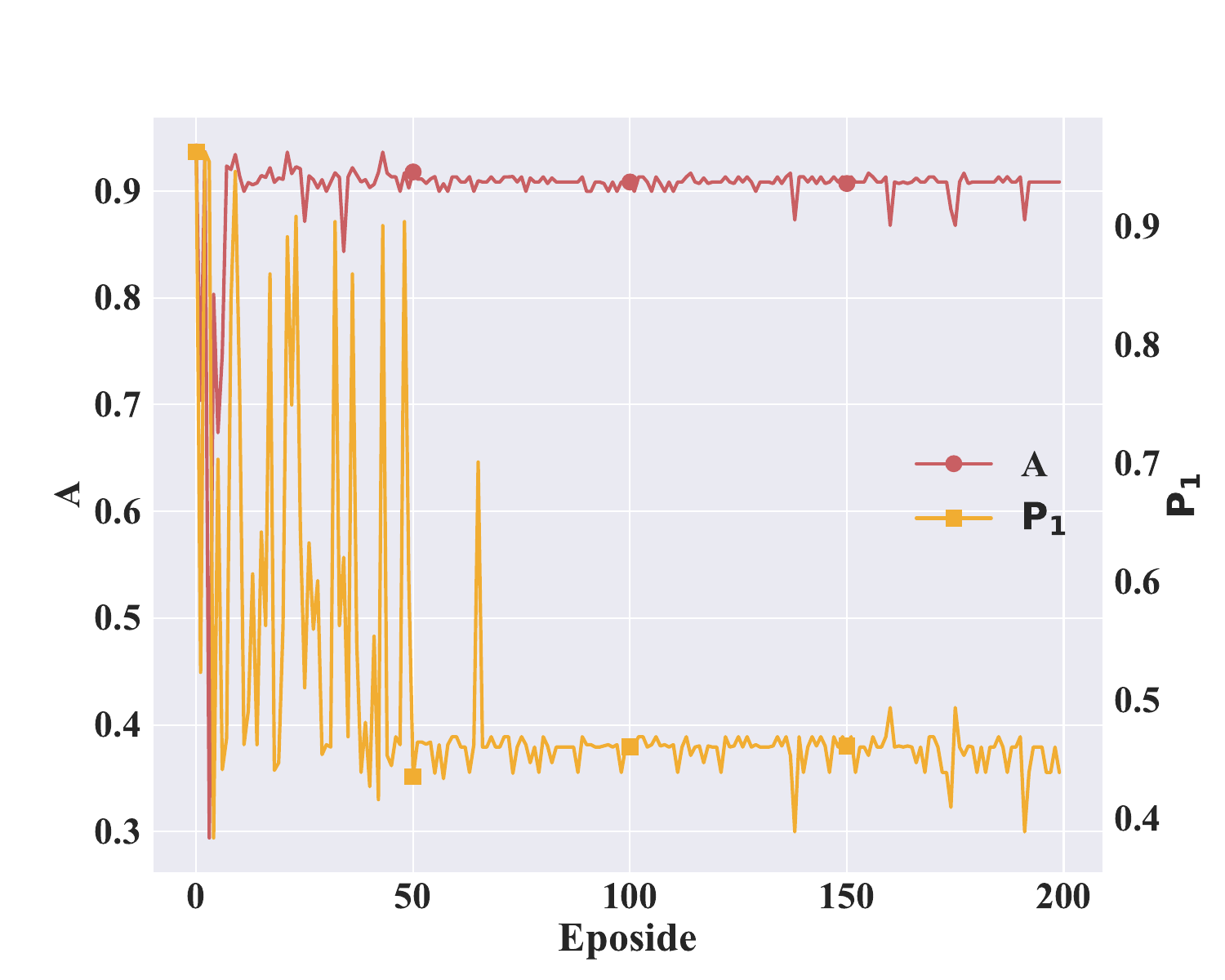}}
	\subfigure[VGG16 Results on CIFAR-10]{
		\label{fig:rl_4}
		\includegraphics[width=0.32\textwidth]{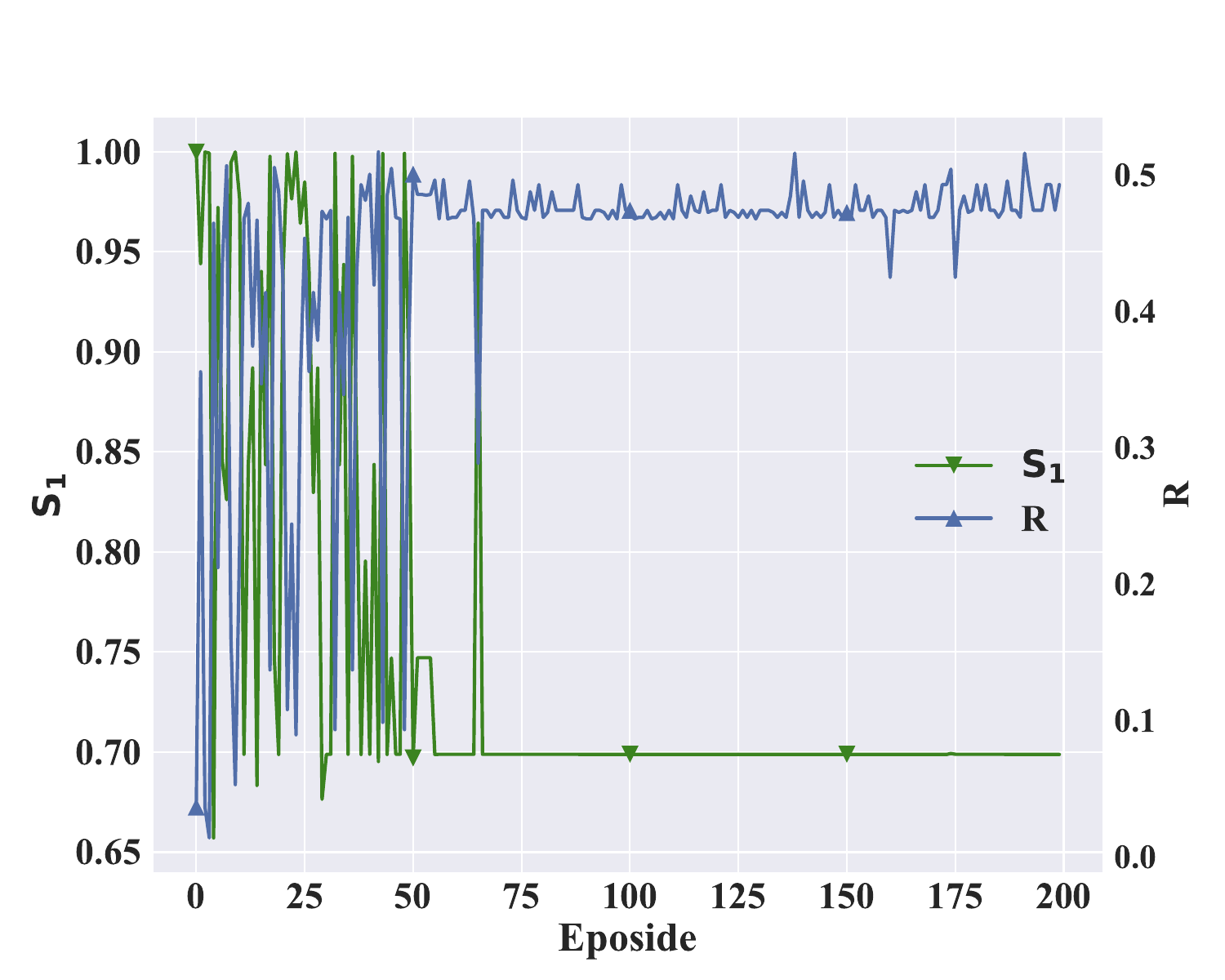}}
	\subfigure[VGG16 Results on CIFAR-100]{
		\label{fig:rl_5}
		\includegraphics[width=0.32\textwidth]{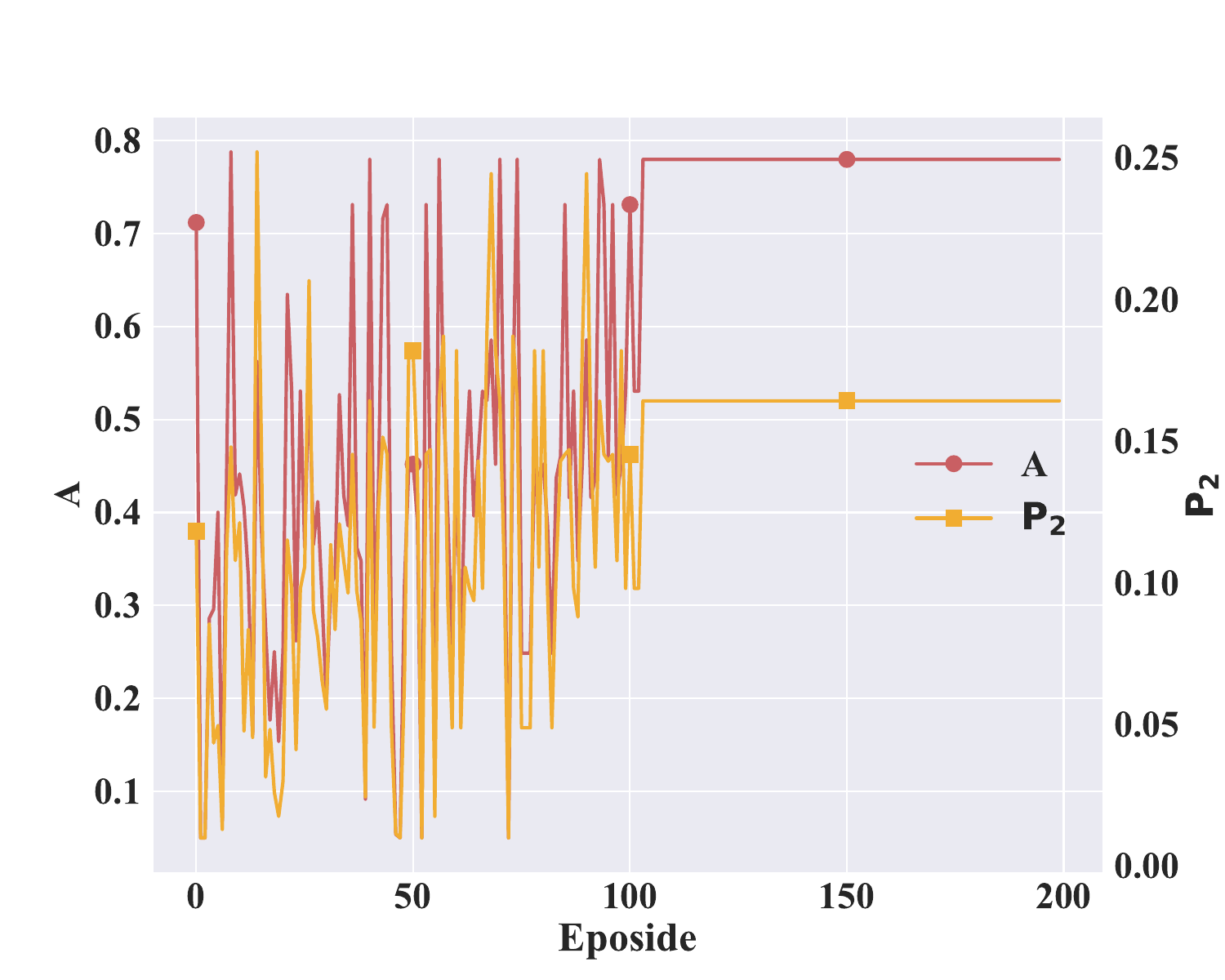}}
	\subfigure[VGG16 Results on CIFAR-100]{
		\label{fig:rl_6}
		\includegraphics[width=0.32\textwidth]{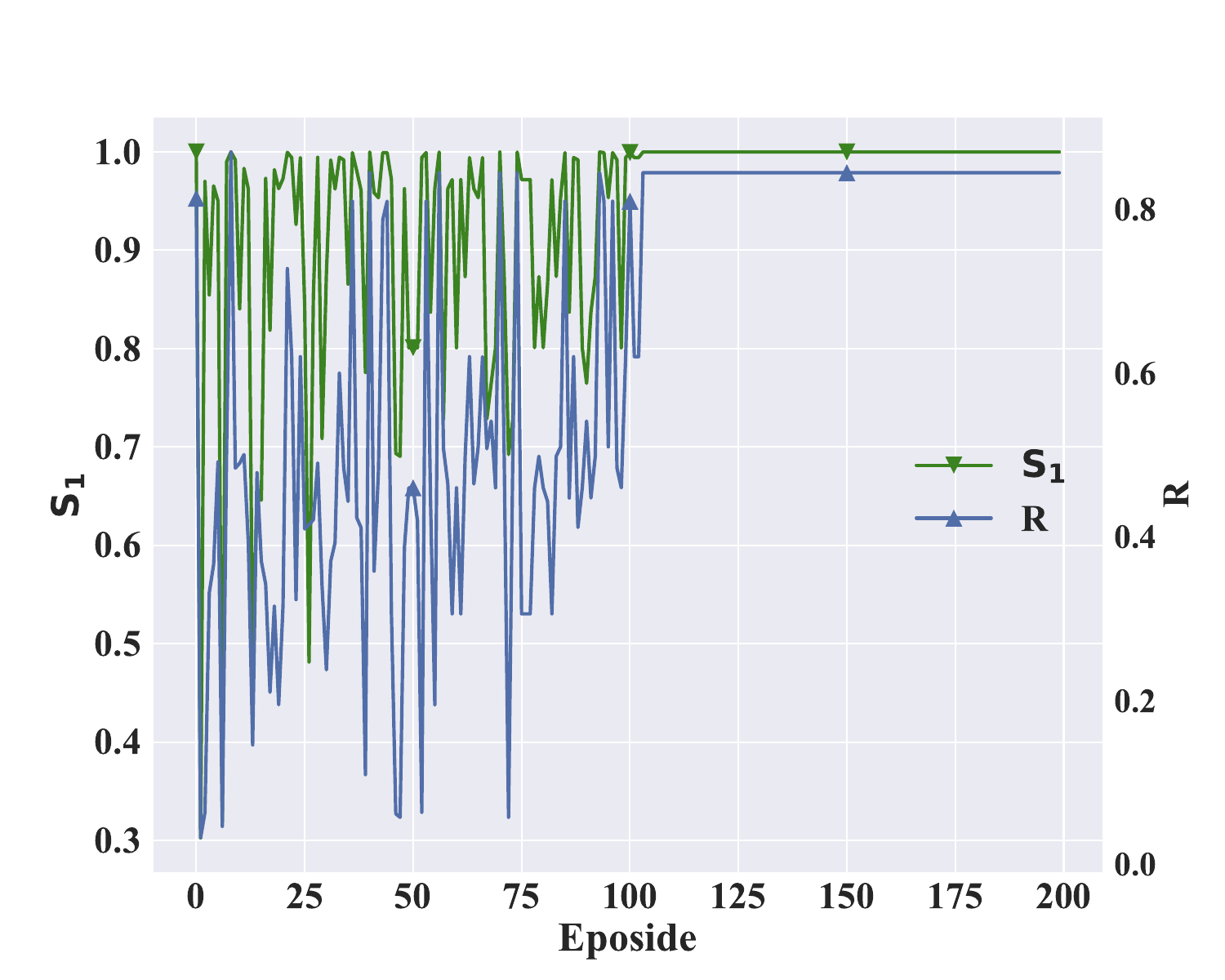}}		
	\caption{The reinforcement learning process under a variety settings: (a)(b) show the results of VGG-11 on CIFAR-10 with the reactive decoder. (c)(d) show the results of VGG-16 on CIFAR-10 with the proactive decoder. (e)(f) show the results of VGG-16 CIFAR-100 with the proactive decoder.
		\label{fig:rl}}
\end{figure*}

Fig.~\ref{fig:rl} show the convergence of our RL-based optimizer. We follow the notations in Sec.~\ref{eqn:formulate} when presenting the results. From Fig.~\ref{fig:rl}(a)(c)(e), we can tell the overall trend is that, the model accuracy remains stable throughout the learning process, while the SSIM and the decoder accuracy gradually decrease indicating privacy level increase. That essentially suggests our optimizer learns to select neural network structures and features which preserves input privacy without degrading the model performance. Fig.~\ref{fig:rl}(b)(d)(f) show that the rewards increase with episodes until convergence.

The results of our RL-based optimizer are given in Tab.~\ref{tab:rl}. %We evaluate the performance over two privacy attacks: for the feature inversion attack, we implement both the reactive and proactive decoders on CIFAR-10. For the property inference attack, we implement proactive decoder on CIFAR-100. 
Our new DNN models only have slight accuracy degradation from the base DNN whereas the resource indicators remain high representing a light computational overhead on mobile devices. Since it is not straightforward to tell the privacy preservation effect and the total reward in the form of raw numbers, we compare the results with baselines in the following.

\begin{table*}[!htbp]
	\centering
	\scriptsize
	\caption{Results of running reinforcement Learning based optimizer over different models and datasets. $A_{base}$ is the base DNN accuracy. $A$, $P_1(P_2)$ and $S$ represent accuracy, privacy loss and resource utilization respectively as defined in Sec.~\ref{eqn:formulate}. $R$ is calculated according to Eq.\ref{eqn:reward}. The default resource utilization metric is $S_1$ and privacy metric is $P_1$ with special cases marked in brackets.} 
	\scalebox{0.95}{\begin{tabular}{c | c  c  c  c  c  c  c  c }
		\toprule
		~ & Model & Partition Layer & Compression Techniques & $A_{\textrm{base}}$ & $A$ & $P_1$($P_2$) & $S_1$($S_2$) & $R$ \\ \hline
		\multirow{12}*{$\begin{gathered} \textrm{CIFAR-10} \\ \textrm{Reactive}\end{gathered}$}
		& VGG11 & 28 & $\begin{gathered} \textrm{0:W2 4:W2 8:W2 11:W2} \\ \textrm{15:W2 18:W2 22:W2 25:W2}\end{gathered}$ & 0.9241 & 0.8959 & 0.4363 & 0.699 & 0.497 \\ \cline{2-9}
		&VGG13 & 29 & 3:W2 7:W2 14:W1 21:W2 24:C1 28:C1 & 0.9423 & 0.911 & 0.4172 & 0.8122 & 0.5435 \\ \cline{2-9}
		&VGG16 & 33 & $\begin{gathered} \textrm{0:W1 3:W1 7:W1 10:W1 14:W1 17:W1} \\ \textrm{20:W1 24:W1 27:W1 30:W1}\end{gathered}$ & 0.9397 & 0.9215 & 0.4096 & 0.8269 & 0.5616 \\ \cline{2-9} 
		& AlexNet & 9 & none & 0.8679 & 0.8673 & 0.3798 & 0.9286 & 0.6166  \\ \cline{2-9}
		& AlexNet ($S_2$) & 13 & 0:C3 8:C3 10:C3 & 0.8679 & 0.8324 & 0.4086 & 0.6671 & 0.5043  \\ \cline{2-9}
		& AlexNet ($S_3$) & 13 & 0:C3 8:C3 10:C3 & 0.8679 & 0.8324 & 0.4086 & 0.5418 & 0.4481  \\ \cline{2-9}
		&Lenet & 5 & 3:C3 & 0.7522 & 0.7502 & 0.5300 & 0.9537 & 0.4678 \\ \cline{2-9}
		&Lenet ($S_2$) & 2 & none & 0.7522 & 0.7314 & 0.6547 & 0.6547 & 0.2374 \\
		\cline{2-9}
		& MobileNet v2 & 0 & none & 0.9279 & 0.9364 & 0.5276 & 0.9996 & 0.4803 \\ \cline{2-9}
		& ResNet50 & 11 & 4:W2 5:W2 6:W2 7:C2 8:C2 10:C2 11:W2 & 0.9438 & 0.9197 & 0.3813 & 0.8978 & 0.5966 \\ \cline{2-9}
		& Inception v3 & 7 & none & 0.9541 & 0.9533 & 0.4013 & 0.6924 & 0.5416 \\ \cline{2-9}
		& Wide Resnet28-10 & 5 & 0:W1 2:W1 5:W1 & 0.9644 & 0.9549 & 0.3961 & 0.949 & 0.5964 \\ \hline
		\multirow{3}*{$\begin{gathered} \textrm{CIFAR-10} \\ \textrm{Proactive}\end{gathered}$}
		& VGG11 & 29 & 8:W2 15:W1 18:W1 22:C3 25:C3 & 0.9241 & 0.904 & 0.3969 & 0.7331 & 0.5479 \\ \cline{2-9}
		& VGG13 & 35 & $\begin{gathered} \textrm{3:W1 7:W1 10:W1 14:W1} \\ \textrm{21:W1 24:C3 28:W1 31:C3}\end{gathered}$ & 0.9423 & 0.9133 & 0.3892 & 0.7779 & 0.5628 \\ \cline{2-9}
		&VGG16 & 38 & $\begin{gathered} \textrm{7:W2 10:W2 17:W1 20:W1} \\ \textrm{24:C3 27:C3 34:W2 37:C1}\end{gathered}$ & 0.9397 & 0.9054 & 0.3854 & 0.8356 & 0.5762 \\ \cline{2-9}
		& AlexNet & 13 & 6:W1 8:W1 10:W2 & 0.8679 & 0.8594 & 0.3799 & 0.9654 & 0.6133 \\ \cline{2-9}
		& LeNet & 3 & 3:W1 & 0.7522 & 0.7334 & 0.5113 & 0.9801 & 0.4763 \\
		\hline
		\multirow{3}*{$\begin{gathered}\textrm{CIFAR-100} \\\textrm{Proactive}\end{gathered}$}
		& VGG11 ($P_2$) & 3 & none & 0.7557 & 0.7176 & 0.1706 & 0.9998 & 0.7876 \\ \cline{2-9}
		&VGG13 ($P_2$) & 1 & none & 0.7673 & 0.7625 & 0.2403 & 0.9998 &  0.7549 \\ \cline{2-9}
		&VGG16($P_2$) & 2 & none & 0.7711 & 0.7797 & 0.1643 & 0.9999 & 0.8450 \\ \hline
		\multirow{1}*{Purchase Reactive}
		& Linear model ($P_2$) & 2 & none & 0.8601 & 0.861 & 0.1764 & 0.472 & 0.7218 \\ \hline
		\multirow{1}*{Tiny ImageNet Reactive}
		& ResNet18 & 11 & $\begin{gathered} \textrm{0:W2 3:W2 4:W2 5:W2 6:W2} \\ \textrm{7:W2 8:W2 9:W2 10:W2}\end{gathered}$ & 0.6806 & 0.658 & 0.4105 & 0.7014 & 0.5191 \\
		\bottomrule
	\end{tabular}}
	\label{tab:rl}
\end{table*}

\begin{table*}[!htbp]
	\centering
	\scriptsize
	\caption{Results of exhaustive search over different models and datasets. $\Delta A$, $\Delta P$, $\Delta S$ and $\Delta R$ are gain of our RL framework in accuracy, privacy loss (smaller is better), resource utilization and reward over the exhaustive search method.}
	\scalebox{0.90}{\begin{tabular}{c | c  c  c | c  c | c  c | c  c | c  c }
		\toprule
		~ & Model & $\begin{gathered}\textrm{Partition}\\\textrm{Layer}\end{gathered}$ & $\begin{gathered}\textrm{Compression}\\\textrm{Techniques}\end{gathered}$ & $A$ & $\Delta A$ & $P_1$ ($P_2$) & $\Delta P$ & $S_1$($S_2$) & $\Delta S$ & $R$ & $\Delta R$ \\ \hline
		\multirow{4}*{$\begin{gathered} \textrm{CIFAR-10} \\ \textrm{Reactive}\end{gathered}$}
		& VGG11 & 21 & 11:C1 18:C1 & 0.8950 & {\bf 0.10\%} & 0.5300 & {\bf -17.68\%} & 0.7950 & -12.08\% & 0.4361 & {\bf 13.96\%} \\ \cline{2-12}
		&VGG13 & 30 & $\begin{gathered} \textrm{0:W1 3:W1 7:W1 10:W1 14:W1} \\ 
		\textrm{17:W1 21:W1 24:W1 28:W1}
		\end{gathered}$ & 0.9353 & -2.6\% & 0.4540 & {\bf -8.11\%} & 0.7502 & {\bf 8.26\%} & 0.5082 & {\bf 6.95\%} \\ \cline{2-12}
		&VGG16 & 33 & $\begin{gathered}
		\textrm{0:W1 3:W1 7:W1 10:W1 14:W1 17:W1} \\
		\textrm{20:W1 24:W1 27:W1 30:W1}
		\end{gathered}$ & 0.9282 & -0.72\% & 0.4527 & {\bf -9.52\%} & 0.92 & -10.12\% & 0.5372 & {\bf 4.54\%} \\ \cline{2-12}
		&AlexNet & 16 & 0:W1 3:W1 6:W1 8:W1 10:W1 14:F1 & 0.8425 & {\bf 2.94\%} & 0.3855 & {\bf -1.48\%} & 0.9587 & -3.14\% & 0.5955 & {\bf 3.54\%} \\ \cline{2-12}
		&LeNet & 7 & 6:F1 & 0.7189 & {\bf 4.35\%} & 0.5196 & 2\% & 0.8176 & {\bf 16.65\%} & 0.4438 & {\bf 5.41\%} \\ 
		\hline
		\multirow{3}*{$\begin{gathered} \textrm{CIFAR-10} \\ \textrm{Proactive}\end{gathered}$}
		& VGG11 & 14 & 0:C3 4:C3 8:C3 11:C3 & 0.8227 & {\bf 9.88\%} & 0.4097 & {\bf -3.12\%} & 0.8609 & -14.84\% & 0.5153 & {\bf 6.33\%} \\ \cline{2-12}
		&VGG13 & 20 & 0:C3 3:C3 7:C3 10:C3 14:C3 17:C3 & 0.7974 & {\bf 14.53\%} & 0.3823 & 1.8\% & 0.8613 & -9.68\% & 0.5126 & {\bf 9.79\%} \\ \cline{2-12}
		&VGG16 & 33 & $\begin{gathered}
		\textrm{0:W1 3:W1 7:W1 10:W1 14:W1}\\
		\textrm{17:W1 20:W1 24:W1 27:W1 30:W1}
		\end{gathered}$ & 0.9090 & -0.40\% & 0.4234 & {\bf -9.0\%} & 0.8269 & {\bf 1.05\%} & 0.5410 & {\bf 6.51\%} \\
		\hline
		\multirow{3}*{$\begin{gathered}\textrm{CIFAR-100} \\\textrm{Proactive}\end{gathered}$}
		& VGG11 ($P_2$) & 21 & 0:W1 4:W1 8:W1 11:W1 15:W1 18:W1 & 0.6889 & {\bf 4.17\%} & 0.1302 & 31.03\% & 0.8372 & {\bf 19.42\%} & 0.7719 & {\bf 2.03\%} \\ \cline{2-12}
		&VGG13 ($P_2$) & 30 & $\begin{gathered}
		\textrm{0:W1 3:W1 7:W1 10:W1 14:W1} \\
		\textrm{17:W1 21:W1 24:W1 28:W1}
		\end{gathered}$ & 0.7717 & -1.19\% & 0.3542 & {\bf -32.16\%} & 0.7502 & {\bf 33.27\%} & 0.6090 & {\bf 23.96\%} \\ \cline{2-12}
		&VGG16 ($P_2$) & 33 & $\begin{gathered}
		\textrm{0:W2 3:W2 7:W2 10:W2 14:W2} \\
		\textrm{17:W2 20:W2 24:W2 27:W2 30:W2}
		\end{gathered}$ & 0.6998 & {\bf 11.42\%} & 0.2405 & {\bf -31.68\%} & 0.8437 & {\bf 18.51\%} & 0.6724 & {\bf 25.67\%} \\
		\bottomrule
	\end{tabular}}
	\label{tab:exhaustive}
\end{table*}

\begin{table*}[!htbp]
	\centering
	\scriptsize
	\caption{Results of random search over different models and datasets. $\Delta A$, $\Delta P$, $\Delta S$ and $\Delta R$ are gain of our RL framework in accuracy, privacy loss (smaller is better), resource utilization and reward over the random search method.}
	\scalebox{0.90}{
	\begin{tabular}{c | c  c  c | c  c | c  c | c  c | c  c }
		\toprule
		~ & Model & $\begin{gathered}\textrm{Partition}\\\textrm{Layer}\end{gathered}$ & $\begin{gathered}\textrm{Compression}\\\textrm{Techniques}\end{gathered}$ & $A$ & $\Delta A$ & $P_1$ ($P_2$) & $\Delta P$ & $S_1$($S_2$) & $\Delta S$ & $R$ & $\Delta R$ \\ \hline
		\multirow{6}*{$\begin{gathered} \textrm{CIFAR-10} \\ \textrm{Reactive}\end{gathered}$}
		& VGG11 & 28 & 11:C3 15:W2 18:C1 22:W1 25:W2 & 0.8641 & {\bf 3.68\%} & 0.4343 & 0.46\% & 0.7226 & -3.25\% & 0.4883 & {\bf 1.78\%}  \\ \cline{2-12}
		& VGG13 & 33 & $\begin{gathered} \textrm{0:W2 7:W2 10:C2 17:W2} \\ \textrm{21:C3 24:C1 28:C1 31:C2} \end{gathered}$  & 0.8611 & {\bf 5.79\%} & 0.4282 & {\bf -2.57\%} & 0.8130 & -0.10\% & 0.5043 & {\bf 7.77\%} \\ \cline{2-12}
		& VGG16 & 36 & $\begin{gathered} \textrm{7:W1 10:W2 17:C3 20:C1} \\ \textrm{24:W1 27:W1 30:C1 34:W2} \end{gathered}$ & 0.8812 & {\bf 4.57\%} & 0.4282 & {\bf -4.34\%} & 0.8166 & {\bf 1.27\%} & 0.5182 & {\bf 8.38\%} \\ \cline{2-12}
		& AlexNet & 13 & 8:W2 & 0.8645 & {\bf 0.32\%} & 0.4076 & {\bf -6.82\%} & 0.9299 & -0.14\% & 0.5872 & {\bf 5.01\%} \\ \cline{2-12}
		& AlexNet($S_2$) & 13 & 0:C3 8:C3 10:W2 & 0.8435 & -1.32\% & 0.4067 & 0.47\% & 0.5334 &  {\bf 25.07\%} & 0.4511 & {\bf 11.79\%} \\ \cline{2-12}
		& AlexNet($S_3$) & 13 & 10:C3 & 0.8629 & -3.53\% & 0.4159 & {\bf -1.76\%} & 0.4641 & {\bf 16.74\%} & 0.414 & {\bf 8.24\%} \\ \cline{2-12}
		& LeNet & 7 & 3:W2 6:F1 & 0.6449 & {\bf 16.33\%} & 0.4472 & 18.52\% & 0.8436 & {\bf 13.05\%} & 0.4624 & {\bf 1.17\%} \\ \cline{2-12}
		& LeNet($S_2$) & 2 & none & 0.7314 & {\bf 0\%} & 0.6547 & {\bf 0\%} & 0.4587 & {\bf 0\%} & 0.2374 & {\bf 0\%} \\
		\hline
		\multirow{1}*{CIFAR-100 Proactive} & VGG16 & 2 & 0:W2 & 0.76 & {\bf 2.59\%} & 0.1968 & {\bf -16.51\%}  & 0.9999 & {\bf 0\%} & 0.7891 & {\bf 7.08\%} \\ \hline
		%\multirow{1}*{Purchase} & Linear model & 4 & 0:F1 3:F2 & 0.8182 & {\bf 5.23\%} & 0.217 & {\bf -18.71\%} & 0.8346 & -43.45\% & 0.7242 & -0.33\% \\ \hline
		\multirow{1}*{Purchase Reactive} & Linear model & 2 & none & 0.861 & {\bf 0\%} & 0.1764 & {\bf 0\%} & 0.472 & {\bf 0\%} & 0.7218 & {\bf 0\%} \\ \hline
		\multirow{1}*{Tiny ImageNet Reactive} & ResNet18 & 11 & 3:C1 4:C1 6:W1 9:W2 10:C1 & 0.6412 & {\bf 2.62\%} & 0.3991 & 2.86\% & 0.633 & {\bf 10.81\%} & 0.4899 & {\bf 5.96\%} \\ 
		\bottomrule
	\end{tabular}}
	\label{tab:random}
\end{table*}

\begin{table*}[!htbp]
	\centering
	\scriptsize
	\caption{Results of applying differential privacy to the models: randomized Gaussian noises are inserted to the selected partition layers. $\Delta A$, $\Delta P$, $\Delta S$ and $\Delta R$ are gain of our RL framework in accuracy, privacy loss (smaller is better), resource utilization and reward over differential privacy.}
	\scalebox{1.0}{\begin{tabular}{c | c | c | c | c  c | c  c | c  c | c  c }
		\toprule
		~ & Model & Partition Layer & STD & $A$ & $\Delta A$ & $P_1$ ($P_2$) & $\Delta P$ & $S_1$($S_2$) & $\Delta S$ & $R$ & $\Delta R$ \\ \hline
		\multirow{3}*{$\begin{gathered} \textrm{CIFAR-10} \\ \textrm{Reactive}\end{gathered}$}
		& VGG11 & 15 & 2.0$\times$ & 0.8747 & {\bf 2.42\% } &	0.4700 & {\bf -7.17\% } & 0.7679 & -8.97\% & 0.4747 & {\bf 4.70\% }  \\ \cline{2-12}
		& VGG13 & 20 & 4.0$\times$ & 0.8746 & {\bf 4.16\% } & 0.4860 & {\bf -14.16\%  }& 0.8782 & -7.52\% & 0.4700 & {\bf 15.64\% } \\ \cline{2-12}
		& VGG16 & 24 & 5.0$\times$ & 0.9042 & {\bf 1.91\% } & 0.4291 & {\bf -4.54\% } & 0.8019 & {\bf 3.12\% } & 0.5277 & {\bf 6.42\% } \\ \hline
		\multirow{3}*{$\begin{gathered} \textrm{CIFAR-10} \\ \textrm{Proactive}\end{gathered}$}
		& VGG11 & 15 & 2.0$\times$ & 0.8747 & {\bf 3.35\% } & 0.4700 & {\bf -15.55\% } & 0.7679 & -4.53\% & 0.4747 & {\bf 15.42\% } \\ \cline{2-12}
		& VGG13 & 21 & 4.0$\times$ & 0.8858 & {\bf 3.1\% } & 0.4832 & {\bf -19.45\% } & 0.7289 & {\bf 6.72\% } & 0.4501 & {\bf 25.04\% } \\ \cline{2-12}
		& VGG16 & 24 & 5.0$\times$ & 0.9042 & {\bf 0.13\% } & 0.4291 & {\bf -10.21\% } & 0.8019 & {\bf 4.20\% } & 0.5277 & {\bf 9.19\% } \\
		\bottomrule
	\end{tabular}}
	\label{tab:noise}
\end{table*}

\begin{table*}[!htbp]
	\centering
	\scriptsize
	\caption{Results of applying adversarial learning to models with CIFAR-10. $\Delta A$, $\Delta P$, $\Delta S$ and $\Delta R$ are gain of our framework in accuracy, privacy loss (smaller is better), resource utilization and reward over adversarial learning.}
	\begin{tabular}{ c | c | c  c | c  c | c  c | c  c }
		\toprule
		Model & Partition Layer & A & $\Delta A$ & P & $\Delta P$ & S & $\Delta S$ & R & $\Delta R$ \\ \hline
		VGG11 & 15 & 0.8798 & {\bf 2.75\% } & 0.6913 & {\bf -42.59\% } & 0.7679 & -4.53\% & 0.2780 & {\bf 97.07\% } \\  \hline
		VGG13 & 21 & 0.9029 & {\bf 1.15\% } & 0.7289 & {\bf -46.6\% } & 0.7528 & {\bf 3.33\% } & 0.2439 & {\bf 130.75\% }\\ \hline
		VGG16 & 24 & 0.9111 & -0.63\% & 0.7595 & {\bf -49.27\% } & 0.8019 & {\bf 4.20\% } & 0.2241 & {\bf 157.12\% } \\ 
		\bottomrule
	\end{tabular}
	\label{tab:adv}
\end{table*}

\subsubsection{Baselines}
We compare our framework with the following baselines. The first two are exhaustive search and random search to compare the compression and partitioning strategies. To verify that our approach is effective in preserving privacy, we compare it against two other privacy-preserving methods: differential privacy and adversarial learning. The detail of each baseline is introduced as below.

{\bf Comparison with Exhaustive Search.} Since exhaustively searching over all combinations of compression and partition strategies is almost impossible due to large search space, we narrow down the search space to the combination of three representative partitioning positions, {\em i.e,} front, middle and rear, and only one type compression technique selected from Tab.~\ref{tab:com}. Two adversarial retraining strategies are applied to the searched models accordingly. The results of exhaustive search with the highest reward are given in Tab.~\ref{tab:exhaustive}. Our RL-based optimizer outperforms exhaustive search in almost all cases, with an average improvement of $9.88\%$ in reward value. On average, the model accuracies achieve $3.86\%$ enhancement (with moderate decreases in some cases), the privacy loss reduces by $7.08\%$, and the resource utilization increases by $4.3\%$ compared to exhaustive search. The results demonstrate the effectiveness of RL-based optimizer in seeking neural network structures and placement to obtain an optimal tradeoff among accuracy, privacy, and resource utilization.

{\bf Comparison with Random Search.} We compare our RL-based optimizer with a random search strategy which randomly selects the combination of partition layer and compression techniques. We tried 100 different combinations and show the results with the highest rewards in Tab.~\ref{tab:random}. Our RL-based optimizer outperforms random search in almost all cases, with $5.2\%$ average improvement in reward, $2.82\%$ average improvements in accuracy, $0.88\%$ average decrement in privacy loss, and $5.77\%$ average increase in resource utilization. For LeNet($S_2$) on CIFAR-10 Reactive and Linear model on Purchase, we observe that the same partition layer and compression techniques are chosen for both random search and our method, since the search space for the two models are restricted. %For LeNet on CIFAR-10 Reactive, although random search achieves higher reward, the accuracy drop is unacceptable. 
Overall, our method achieves better tradeoffs.

{\bf Comparison with Differential Privacy.} We compare our method with the differential privacy (DP) mechanism which inserts additional Gaussian noise to the intermediate-layer features to preserve privacy. RL-based optimizer automatically chooses the partition layer and compression techniques while DP manually chooses the partition layer. For a fair comparison on reward, we select the partition layer which leads to a similar resource utilization with our RL-based scheme and inserts the noise to that layer. The randomized noise is drawn from normal distributions with standard deviation (STD) $0.1, 0.5, 1, 2, 3, 4, 5, 10$ times of the feature STD respectively. We observe noise with a higher STD generally leads to a higher privacy level but unfortunately a lower accuracy. For conciseness, we only report results with the highest reward for different noise STDs in Tab.~\ref{tab:noise}. Both reactive decoder and the proactive decoder are chosen in reward calculation. On average, our RL-based approach exceeds differential privacy by $12.74\%$ in reward, with a reduction of $11.85\%$ in privacy loss and a mild improvement of $2.51\%$ in accuracy. The result confirms that modifying the neural network structure can achieve a higher level of privacy compared with feature perturbations.

{\bf Comparison with Adversarial learning.} \cite{xiao2019adversarial} introduces an adversarial learning framework that prevents an adversary from decoding the latent representations of the original input. Without considering the neural network structure, the approach maximizes reconstruction error over the encoder weights. %Adversarial learning \cite{xiao2019adversarial} is adopted to adjust the model weights to fight against privacy leakage while maintaining accuracy on the neural network. 
We compare our proactive framework with adversarial learning \cite{xiao2019adversarial} to show the power of neural architecture search for privacy preservation. The results with the highest reward are shown in Tab.~\ref{tab:adv}. We select the partition layer which leads to a similar resource utilization with the RL-based approach. Results show that the RL-based method achieves $128.33\%$ enhancement in reward, $1.09\%$ improvement in accuracy and $46.15\%$ reduction in privacy loss. Hence evidence have shown that modifying neural network architecture is indeed helpful for finding a better tradeoff.

\subsection{Transferability}

It is time-consuming if every time we apply our framework to a new base DNN, the policy network has to be trained from scratch. Hence we investigate the transferability of the policy across different models and datasets.

\textbf{Transfer across different base DNNs.} Base DNN model determines the state of the RL optimizer, i.e., the input to the policy network. To see if our trained policy network for one base DNN transfers to another, we first train the optimizer on VGG11, and then apply the policy network to other base DNNs. Since a pre-trained policy network and a randomly initialized policy network both eventually converge to the optimal policy given a sufficient number of training episodes, we compare the performance of the searched neural networks at the $10^{th}$ episode. In addition, we present the number of episodes $N$ until convergence.
\begin{table*}[!htbp]
	\centering
	\scriptsize
	\caption{Results of transfer learning. The table compares the $10^{th}$ episode-performance of the searched model from the pre-trained policy network and the train-from-scratch policy network. `Episode' denotes the number of episodes until convergence. }
	\scalebox{0.85}{
		\begin{tabular}{c | c  c  c  c  c | c  c  c  c  c | c  c  c  c  c}
			\toprule
			\multirow{2}{*}{~} & \multicolumn{5}{ c | }{VGG11 $\rightarrow$ VGG13} & \multicolumn{5}{ c | }{VGG11 $\rightarrow$ VGG16}& \multicolumn{5}{ c }{VGG11 $\rightarrow$ AlexNet} \\ \cline{2-16}
			& R & A & P & S & Episode & R & A & P & S & Episode & R & A & P & S & Episode \\ \hline
			Pre-trained & {\bf 0.4888 } & {\bf 0.9249 } & 0.4879 & {\bf 0.8338 } & 77 & {\bf 0.5091 } & {\bf 0.9201 } & {\bf 0.4150 } & 0.6665 & 33 & {\bf 0.5045 } & {\bf 0.7078 } & {\bf 0.3808 } & 0.9678 & 61 \\ \hline
			Scratch & 0.1502 & 0.5609 & {\bf 0.4499 } & 0.2642 & 161 & 0.1969 & 0.3533 & 0.4583 & {\bf 0.8185 } & 159 & 0.2940 & 0.4127 & 0.3811 & {\bf 0.9710 } & 96 \\ 
			\bottomrule
		\end{tabular}
		\label{tab:transfer_learning_model}}
\end{table*}

Tab.~\ref{tab:transfer_learning_model} displays the transfer results. There is a slight increase in privacy loss when transferring from VGG11 to VGG13, and a mild degradation in resource utilization when transferring the policy from VGG11 to VGG16 or AlexNet. Above all, at the $10^{th}$ episode, the reward of the pre-trained policy is almost always higher than that trained from scratch, showing evidence of faster convergence in the transferred policy, despite of different DNN structures. Moreover, transferred policy networks require a smaller episode number indicating shorter training time by loading pre-trained policy network.

\textbf{Transfer across different datasets.} The transferability across different datasets is important in the case that the original dataset is huge and is time-consuming to find an optimal deployment strategy. It makes sense to transfer policy from smaller datasets to large ones to reduce training efforts. To verify the point, we train our policy network on ResNet18, CIFAR-10, and then transfer the policy network to ResNet18, Tiny ImageNet. We show the performance of the transferred policy network and the one trained from scratch on Tiny ImageNet in Tab.~\ref{tab:transfer_learning_dataset}. It is observed that pre-trained policy network reaches high reward only after a few episodes and requires fewer episodes to achieve convergence.

\begin{table}[!htbp]
	\centering
	\scriptsize
	\caption{Results of transfer learning. The table compares the $10^{th}$ episode-performance of the searched model from the pre-trained policy network and the train-from-scratch policy network. `Episode' denotes the number of episodes until convergence.}
	\begin{tabular}{c | c  c  c  c  c }
		\toprule
		\multirow{2}{*}{Resnet18} & \multicolumn{5}{ c }{CIFAR-10 $\rightarrow$ Tiny ImageNet} \\ \cline{2-6}
		& R & A & P & S & Episode \\ \hline
		Pre-trained & {\bf 0.4968} & {\bf 0.6606} & {\bf 0.3968} & {\bf 0.6108} & 27 \\ \hline
		Scratch & 0.1393 & 0.5988 & 0.8390 & 0.8720 & 99  \\ 
		\bottomrule
	\end{tabular}
	\label{tab:transfer_learning_dataset}
\end{table}

\begin{table}[!htbp]
	\centering
	\scriptsize
	\caption{Nvidia TX2 results for AlexNet CIFAR-10.}
	\scalebox{1.0}{
		\begin{tabular}{ c | c | c | c | c }
			\toprule
			AlexNet & Params($\times 10^6$) & MAC($\times 10^6$) & Latency(s) & Reward \\ \hline
			RL($S_3$) & {\bf 0.95} & {\bf 20.89} & {\bf 4.78} &  {\bf 0.4481}    \\ \hline
			Random Search($S_3$) & 1.73   & 33.3  & 5.09  &  0.414       \\ \hline
			RL($S_2$) & {\bf 0.95} & {\bf 20.9} & {\bf 4.81} & {\bf 0.5043}   \\ \hline
			Random Search($S_2$) & 1.06  & 29.3 & 5.16 & 0.4511 \\
			\bottomrule
	\end{tabular}}
	\label{tab:tx2}
\end{table}

\begin{table}[!htbp]
	\centering
	\scriptsize
		\caption{Results of customized criteria on CIFAR-10 with reactive adversary. $P$ denotes the privacy loss which should be less than the defined threshold.}
		\begin{tabular}{ c |c  c  c | c  c  c}
			\toprule
			CIFAR-10& $A_m$ & $P_m$ & $S_m$ & A & $P_1$ & $S_1$ \\ \hline
			& 0 & 0 & 0 &  0.8959 &  0.4363 & 0.699 \\ \cline{2-7} 
			\multirow{-2}{*}{VGG11} & 0.85 & 0 & 0.8 & 0.8526 & 0.541 & 0.874  \\ \hline
			& 0 & 0 & 0 & 0.7505 & 0.5808 & 0.9537 \\ \cline{2-7} 
			\multirow{-2}{*}{Lenet} & 0 & 0.5 & 0 & 0.6775 & 0.4274 & 0.7866 \\ 
			\bottomrule
	\end{tabular}
	\label{tab:criteria}
\end{table}

\subsection{Real-World Performance}

We deploy the searched model by our algorithm on {\tt Nvidia TX2} to see how it performs in real-world settings. With AlexNet as the base DNN, the results are displayed in Tab.~\ref{tab:tx2}. We present the actual parameters, MACs of encoder, and latency including $T_e$ and $T_t$. We observe that our RL algorithm enjoys lower resource costs and higher reward than the random search baseline.

\textbf{Customized Criteria.}
In Eq.~\ref{eqn:reward}, accuracy, privacy and resource utilization are treated equally. In real-world settings, users could have different weights on the three metrics and hence the optimal partition and compression actions would be different. We evaluate the case where users have customized criteria which are expressed as:

\begin{equation}
	\label{eqn:reward_variant}
	R = R_{A} \times R_{P} \times R_{S} \times 1\{A>A_{m}, P<P_{m}, S>S_{m}\}
\end{equation}
where $A_{m}$, $P_{m}$ and $S_{m}$ are the tolerance threshold for accuracy, privacy loss and resource utilization respectively. Tab.~\ref{tab:criteria} gives results for VGG11 and LeNet under different criteria. As we can tell, if specific thresholds are defined, our policy network would seek the optimal model satisfying the constraints.

\section{Conclusion}
Realizing that the neural network features pose significant threats to input data privacy, we propose to apply partition and compression to the neural network to obtain privacy-preserving mobile-cloud configurations. To achieve a better tradeoff among the objectives of privacy, accuracy and resource utilization, we present a reinforcement learning based optimizer searching for the optimal neural network transformation and placement strategies across the mobile device and the cloud. As the experimental results show, our RL-based optimizer has superior performance in preserving input privacy by tapping into DNN structural flexibility, with little degradation of the model performance. Moreover, the policy trained on one DNN can be easily transferred to another, facilitating the RL-based searching.

\bibliographystyle{unsrt}  
\bibliography{main}  %%% Remove comment to use the external .bib file (using bibtex).
%%% and comment out the ``thebibliography'' section.

%%% Comment out this section when you \bibliography{references} is enabled.
%\begin{thebibliography}{1}
%
%\bibitem{kour2014real}
%George Kour and Raid Saabne.
%\newblock Real-time segmentation of on-line handwritten arabic script.
%\newblock In {\em Frontiers in Handwriting Recognition (ICFHR), 2014 14th
%  International Conference on}, pages 417--422. IEEE, 2014.
%
%\bibitem{kour2014fast}
%George Kour and Raid Saabne.
%\newblock Fast classification of handwritten on-line arabic characters.
%\newblock In {\em Soft Computing and Pattern Recognition (SoCPaR), 2014 6th
%  International Conference of}, pages 312--318. IEEE, 2014.
%
%\bibitem{hadash2018estimate}
%Guy Hadash, Einat Kermany, Boaz Carmeli, Ofer Lavi, George Kour, and Alon
%  Jacovi.
%\newblock Estimate and replace: A novel approach to integrating deep neural
%  networks with existing applications.
%\newblock {\em arXiv preprint arXiv:1804.09028}, 2018.
%
%\end{thebibliography}

\end{document}